\documentclass[11pt,a4paper]{article}
\usepackage[hyperref]{acl2020}
\usepackage{times}
\usepackage{latexsym}
\usepackage{url}
\usepackage[T5,T1]{fontenc}
\usepackage[utf8]{inputenc}
\usepackage{booktabs}
\usepackage{multirow}
\usepackage{amssymb}
\usepackage{amsmath}
\usepackage{algorithm}
\usepackage[noend]{algpseudocode}
\usepackage{graphicx}
\usepackage{bbm}
\usepackage{lipsum}
\usepackage{adjustbox}
\usepackage{array}
\usepackage{xcolor}
\usepackage{xcolor,colortbl}
\usepackage{subcaption}
\DeclareCaptionSubType*[Alph]{figure}
\usepackage{enumitem}
\setlist[enumerate]{itemsep=0mm}

\newcolumntype{A}{>{\centering\arraybackslash}m{0.52in}}
\newcolumntype{B}{>{\centering\arraybackslash}m{0.78in}}
\newcolumntype{C}{>{\centering\arraybackslash}m{0.45in}}

\usepackage{tikz}
\usepackage{pgfplots}
\pgfplotsset{compat=newest}
\usepgfplotslibrary{groupplots}

\aclfinalcopy 

\DeclareTextSymbolDefault{\OHORN}{T5}
\DeclareTextSymbolDefault{\UHORN}{T5}
\DeclareTextSymbolDefault{\ohorn}{T5}
\DeclareTextSymbolDefault{\uhorn}{T5}

\title{Predicting Performance for Natural Language Processing Tasks}

\author{Mengzhou Xia, Antonios Anastasopoulos, Ruochen Xu, Yiming Yang, Graham Neubig \\
  Language Technologies Institute, Carnegie Mellon University \\
  \texttt{\{mengzhox,aanastas,yiming,gneubig\}@cs.cmu.edu} \\ \texttt{ruochenx@gmail.com}
  }

\date{}

\begin{document}
\maketitle
\begin{abstract}
Given the complexity of combinations of tasks, languages, and domains in natural language processing (NLP) research, it is computationally prohibitive to exhaustively test newly proposed models on each possible experimental setting. In this work, we attempt to explore the possibility of gaining plausible judgments of how well an NLP model can perform under an experimental setting, \textit{without actually training or testing the model}. To do so, we build regression models to predict the evaluation score of an NLP experiment given the experimental settings as input. Experimenting on~9 different NLP tasks, we find that our predictors can produce meaningful predictions over unseen languages and different modeling architectures, outperforming reasonable baselines as well as human experts. 
Going further, we outline how our predictor can be used to find a small subset of representative experiments that should be run in order to obtain plausible predictions for all other experimental settings.\footnote{Code, data and logs are publicly available at \url{https://github.com/xiamengzhou/NLPerf}.}


\end{abstract}

\section{Introduction}
\label{sec:intro}
\begin{table*}[t]
    \centering
    \small
    \newcommand\redq{\textcolor{red}{?}}
    \begin{tabular}{@{}c|c@{}c@{}c@{}c@{}c@{}c@{}c@{}c@{}c@{}c@{}c@{}c@{}}
    \toprule
        \multirow{2}{*}{BLI Method} & \multicolumn{11}{c}{Evaluation Set}  \\
         & \ \small\textsc{de--en} \ & \ \small\textsc{en--de} \ & \ \small\textsc{es--en} \ & \ \small\textsc{en--es} \ & \ \small\textsc{fr--en} \ & \ \small\textsc{en--fr} \ & \  \small\textsc{it--en} \ & \ \small\textsc{en--it} \ & \ \small\textsc{en--pt} \ & \  \small\textsc{en--ru} \ & \ \small\textsc{es--de} \ & \ \small\textsc{pt--ru} \\
    \midrule
    \citet{zhang2017earth} & \redq  & \checkmark & \checkmark & \checkmark & \redq & \redq & \checkmark & \redq & \redq & \redq & \redq & \redq \\
    \citet{chen-cardie-2018-unsupervised} & \checkmark & \checkmark & \checkmark & \checkmark & \checkmark & \checkmark & \checkmark & \checkmark & \checkmark & \redq & \checkmark  & \redq \\
    \citet{yang2019maam} & \checkmark & \checkmark & \checkmark & \checkmark & \checkmark & \checkmark & \checkmark  & \redq & \redq & \redq & \redq  & \redq \\
    \citet{heyman2019learning} & \redq & \checkmark & \redq & \checkmark & \redq & \checkmark & \redq & \checkmark & \redq & \redq & \redq  & \redq \\
    \citet{huang-etal-2019-hubless} & \redq & \redq & \checkmark & \checkmark  & \checkmark  & \checkmark  & \redq & \redq & \redq & \redq & \redq & \redq\\
    \citet{artetxe-etal-2019-bilingual}  & \checkmark  & \checkmark & \checkmark  & \checkmark & \checkmark  & \checkmark & \redq & \redq & \redq & \checkmark & \redq & \redq\\
    \bottomrule
    \end{tabular}
    \caption{An illustration of the comparability issues across methods and multiple evaluation datasets from the Bilingual Lexicon Induction task. Our prediction model can reasonably fill in the blanks, as illustrated in Section~\ref{sec:perform}.}
    \label{tab:example}
\end{table*}
Natural language processing (NLP) is an extraordinarily vast field, with a wide variety of models being applied to a multitude of tasks across a plenitude of domains and languages. In order to measure progress in all these scenarios, it is necessary to compare performance on test datasets representing each scenario. However, the cross-product of tasks, languages, and domains creates an explosion of potential application scenarios, and it is infeasible to collect high-quality test sets for each. In addition, even for tasks where we do have a wide variety of test data, e.g.~for well-resourced tasks such as machine translation (MT), it is still computationally prohibitive as well as not environmentally friendly \cite{strubell-etal-2019-energy} to build and test on systems for all languages or domains we are interested in. Because of this, the common practice is to test new methods on a small number of languages or domains, often semi-arbitrarily chosen based on previous work or the experimenters' intuition.

As a result, this practice impedes the NLP community from gaining a comprehensive understanding of newly-proposed models.
\autoref{tab:example} illustrates this fact with an example from bilingual lexicon induction, a task that aims to find word translation pairs from cross-lingual word embeddings. As vividly displayed in \autoref{tab:example}, almost all the works report evaluation results on a different subset of language pairs. Evaluating only on a small subset raises concerns about making inferences when comparing the merits of these methods: there is no guarantee that performance on English--Spanish (\textsc{en--es}, the only common evaluation dataset) is representative of the expected performance of the models over all other language pairs \cite{anastasopoulos20embeddings}. Such phenomena lead us to consider if it is possible to make a decently accurate estimation for the performance over an untested language pair without actually running the NLP model to bypass the computation restriction.

Toward that end, through drawing on the idea of characterizing an experiment from \citet{lin-etal-2019-choosing}, we propose a framework, which we call \textsc{\textbf{NLPerf}}, to provide an exploratory solution. We build regression models, to \emph{predict the performance on a particular experimental setting} given past experimental records of the same task, with each record consisting of a characterization of its training dataset and a performance score of the corresponding metric. Concretely, in \S\ref{sec:formulation}, we start with a partly populated table (such as the one from \autoref{tab:example}) and attempt to infer the missing values with the predictor. We begin by introducing the process of characterizing an NLP experiment for each task in ~\S\ref{sec:featuring}. We evaluate the effectiveness and robustness of \textsc{NLPerf} by comparing to multiple baselines, human experts, and by perturbing a single feature to simulate a grid search over that feature (\S\ref{sec:perform}). Evaluations on multiple tasks show that \textsc{NLPerf} is able to outperform all baselines. Notably, on a machine translation (MT) task, the predictions made by the predictor turn out to be more accurate than human experts.

An effective predictor can be very useful for multiple applications associated with practical scenarios. In ~\S\ref{sec:representativeness}, we show how it is possible to adopt the predictor as a scoring function to find a small subset of experiments that are most \emph{representative} of a bigger set of experiments. We argue that this will allow researchers to make informed decisions on what datasets to use for training and evaluation, in the case where they cannot experiment on all experimental settings. Last, in ~\S\ref{sec:model}, we show that we can adequately predict the performance of new models even with a minimal number of experimental records.




\section{Problem Formulation}
\label{sec:formulation}
In this section we formalize the problem of predicting performance on supervised NLP tasks.
Given an NLP model of architecture $\mathcal{M}$ trained over dataset(s) $\mathcal{D}$ of a specific task involving language(s) $\mathcal{L}$ with a training procedure (optimization algorithms, learning rate scheduling etc.) $\mathcal{P}$, we can test the model on a test dataset $\mathcal{D'}$ and get a score $\mathcal{S}$ of a specific evaluation metric.
The resulting score will surely vary depending on all the above mentioned factors, and we denote this relation as $g$:
\begin{equation}
\mathcal{S}_{{\mathcal{M},\mathcal{P},\mathcal{L},\mathcal{D},\mathcal{D'}}} = g(\mathcal{M}, \mathcal{P}, \mathcal{L}, \mathcal{D}, \mathcal{D'}).
\label{eq:formulation}
\end{equation}
In the ideal scenario, for each test dataset $\mathcal{D'}$ of a specific task, one could enumerate all different settings and find the one that leads to the best performance. As mentioned in Section~\S\ref{sec:intro}, however, such a brute-force method is computationally infeasible. Thus, we turn to modeling the process and formulating our problem as a regression task by using a parametric function $f_\theta$ to approximate the true function $g$ as follows:
\[
\hat{\mathcal{S}}_{{\mathcal{M},\mathcal{P},\mathcal{L},\mathcal{D},\mathcal{D'}}} = f_\theta([\Phi_\mathcal{M};\Phi_\mathcal{P};\Phi_\mathcal{L};\Phi_\mathcal{D};\Phi_\mathcal{D'}])
\label{eq:approx}
\]  
where $\Phi_\mathcal{*}$ denotes a set of features for each influencing factor.

For the purpose of this study, we mainly focus on dataset and language features $\Phi_{\mathcal{L}}$ and $\Phi_{\mathcal{D}}$, as this already results in a significant search space, and gathering extensive experimental results with fine-grained tuning over model and training hyperparameters is both expensive and relatively complicated.
In the cases where we handle multiple models, we only use a single categorical model feature to denote the combination of model architecture and training procedure, denoted as $\Phi_\mathcal{C}$. We still use the term \textit{model} to refer to this combination in the rest of the paper. We also omit the test set features, under the assumption that the data distributions for training and testing data are the same (a fairly reasonable assumption if we ignore possible domain shift). Therefore, for all experiments below, our final prediction function is the following:
\[
\hat{\mathcal{S}}_{\mathcal{C}, \mathcal{L},\mathcal{D}} = f_\theta([\Phi_\mathcal{C};\Phi_\mathcal{L};\Phi_\mathcal{D}])
\]  

In the next section we describe concrete instantiations of this function for several NLP tasks.

\section{NLP Task Instantiations}
\label{sec:featuring}

\begin{table*}[t]
\centering
\begin{tabular}{llcccccc}
\toprule
\multirow{2}{*}{Task} & \multirow{2}{*}{Dataset Citation}  & Source & Target & Transfer & \multirow{2}{*}{\# Models} & \multirow{2}{*}{\# EXs} & Task \\
   &   & Langs & Langs & Langs & &  & Metric \\\midrule
 \small Wiki-MT     & \citet{schwenk2019wikimatrix}  & 39 & 39 & --  & single      & 995                    & BLEU               \\
\small TED-MT      &    \citet{qi18naacl}                                         & 54    & 1 & -- & single  & 54                      & BLEU               \\
\small TSF-MT      &   \citet{qi18naacl} & 54 & 1 & 54 & single & 2862 & BLEU               \\
\small TSF-PARSING &    \citet{11234/1-2837} & -- & 30 & 30 & single      & 870          & Accuracy           \\
\small TSF-POS     &     \citet{11234/1-2837}                                        & -- & 26    & 60 & single     & 1531       & Accuracy           \\
\small TSF-EL      &   \citet{rijhwani19aaai} & -- & 9 & 54        & single   & 477                     & Accuracy           \\
\small BLI          & \citet{lample2018word}   & 44 & 44 & -- & 3  & 88$\times3$                     & Accuracy           \\
\small MA          & \citet{mccarthy-etal-2019-sigmorphon} & -- & 66   & --          & 6  &  107$\times$6                   & F1                 \\
\small UD & \citet{zeman-EtAl:2018:K18-2} &  -- & 53 & -- & 25 & 72$\times25$ & F1 \\
\bottomrule
\end{tabular}
\caption{Statistics of the datasets we use for training predictors. \# EXs denote the total number of experiment instances; Task Metric reflects how the models are evaluated.}
\label{tab:data}
\end{table*}
To build a predictor for NLP task performance, we must 1) select a task, 2) describe its featurization, and 3) train a predictor. We describe details of these three steps in this section.

\paragraph{Tasks} We test on tasks including bilingual lexicon induction (BLI); machine translation trained on aligned Wikipedia data (Wiki-MT), on TED talks (TED-MT), and with cross-lingual transfer for translation into English (TSF-MT); cross-lingual dependency parsing (TSF-Parsing); cross-lingual POS tagging (TSF-POS); cross-lingual entity linking (TSF-EL); morphological analysis (MA) and universal dependency parsing (UD). Basic statistics on the datasets for all tasks are outlined in \autoref{tab:data}.

For Wiki-MT tasks, we collect experimental records directly from the paper describing the corresponding datasets \cite{schwenk2019wikimatrix}. For TED-MT and all the transfer tasks, we use the results of \citet{lin-etal-2019-choosing}. For BLI, we conduct experiments using published results from 
three papers, namely \citet{artetxe2016learning}, \citet{artetxe-etal-2017-learning} and \citet{xu2018unsupervised}.
For MA, we use the results of the SIGMORPHON 2019 shared task~2~\cite{mccarthy-etal-2019-sigmorphon}.
Last, the UD results are taken from the CoNLL 2018 Shared Task on universal dependency parsing~\cite{zeman2018conll}.

\paragraph{Featurization} 
For language features, we utilize six distance features from the URIEL Typological Database \cite{littell2017uriel}, namely geographic, genetic, inventory, syntactic, phonological, and featural distance.

The complete set of dataset features includes the following:
\begin{enumerate}[noitemsep,nolistsep]
    \item \emph{Dataset Size:} The number of data entries used for training.
    \item \emph{Word/Subword Vocabulary Size:} The number of word/subword types.
    \item \emph{Average Sentence Length:} The average length of sentences from all experimental.
    \item \emph{Word/Subword Overlap:} 
    \begin{gather}
        \frac{|T_1 \cap T_2|}{|T_1| + |T_2|} \nonumber
    \end{gather}
    where $T_1$ and $T_2$ denote vocabularies of any two corpora.
    \item \emph{Type-Token Ratio (TTR):} The ratio between the number of types and number of tokens \cite{richards1987type} of one corpus.
    \item \emph{Type-Token Ratio Distance:} 
    \begin{gather}
            \left(1 - \frac{\mathrm{TTR}_1}{\mathrm{TTR}_2}\right)^2 \nonumber
    \end{gather}
    where $\mathrm{TTR}_1$ and $\mathrm{TTR}_2$ denote TTR of any two corpora.
    \item \emph{Single Tag Type:} Number of single tag types.
    \item \emph{Fused Tag Type:} Number of fused tag types.
    \item \emph{Average Tag Length Per Word:} Average number of single tags for each word.
    \item \emph{Dependency Arcs Matching WALS Features:} the proportion of dependency parsing arcs matching the following WALS features, computed over the training set: subject/object/oblique before/after verb and adjective/numeral before/after noun.
\end{enumerate}
For transfer tasks, we use the same set of dataset features ${\Phi}_{\mathcal{D}}$ as \citet{lin-etal-2019-choosing}, including features~1--6 on the source and the transfer language side. We also include language distance features between source and transfer language, as well as between source and target language. For MT tasks, we use features~1--6 and language distance features, but only between the source and target language sides. For MA, we use features~1,~2,~5 and morphological tag related features~7--9. For UD, we use features~1,~2,~5, and~10.
For BLI, 
we use language distance features and URIEL syntactic features for the source language and the target language.



\paragraph{Predictor}
Our prediction model is based on gradient boosting trees~\cite{friedman2001greedy}, implemented with XGBoost~\cite{chen2016xgboost}. This method is widely known as an effective means for solving problems including ranking, classification and regression. 
We also experimented with Gaussian processes~\cite{williams1996gaussian}, but settled on gradient boosted trees because performance was similar and Xgboost's implementation is very efficient through the use of parallelism.
We use squared error as the objective function for the regression and adopted a fixed learning rate~0.1. 
To allow the model to fully fit the data we set the maximum tree depth to be~10 and the number of trees to be~100, and use the default regularization terms to prevent the model from overfitting.


\section{Can We Predict NLP Performance?}
\label{sec:perform}
In this section we investigate the effectiveness of \textsc{NLPerf} across different tasks on various metrics. 
Following \citet{lin-etal-2019-choosing}, we conduct $k$-fold cross validation for evaluation. To be specific, we randomly partition the experimental records of $\langle \mathcal{L}, \mathcal{D}, \mathcal{C}, \mathcal{S} \rangle$ tuples into $k$ folds, and use $k-1$ folds to train a prediction model and evaluate on the remaining fold. Note that this scenario is similar to ``filling in the blanks'' in \autoref{tab:example}, where we have some experimental records that we can train the model on, and predict the remaining ones.

For evaluation, we calculate the average root mean square error (RMSE) between the predicted scores and the true scores.


\begin{table*}[t]
\centering
\resizebox{\textwidth}{!}{%
\setlength\tabcolsep{2.0pt}
\begin{tabular}{!{}l|AAABAACCC!{}}
\toprule
& \multicolumn{9}{c}{Task} \\
Model     & \small Wiki-MT &\small TED-MT &\small TSF-MT & \small TSF-PARSING &\small TSF-POS &\small TSF-EL & \small BLI  &\small MA & \small UD   \\ \midrule
Mean & 6.40    & 12.65  & 10.77  & 17.58  & 29.10  & 18.65  & 20.10 & 9.47 & 17.69   \\
Transfer Lang-wise & -- & -- & 10.96 & 15.68 & 29.98 & 20.55 & -- & -- & -- \\
Source Lang-wise & 5.69 & 12.65 & 2.24 & -- & -- & -- &  20.13 & -- & -- \\
Target Lang-wise & 5.12 & 12.65 & 10.78 & 12.05 & 8.92 & 8.61 & 20.00 & 9.47 & -- \\
 \textsc{NLPerf} (SM) & 2.50 & 6.18 & 1.43 & 6.24 & 7.37 & 7.82 & 12.63  & 6.48 & 12.06 \\
\midrule
Model-wise  & -- & -- & -- & -- & -- & -- & 8.77 & 5.22 & 4.96 \\
 \textsc{NLPerf} (MM)  & --    & --   &  --  & -- & -- & --   & 6.87 & 3.18 & 3.54  \\
 \bottomrule
\end{tabular}}
\caption{RMSE scores of three baselines and our predictions under the single model and multi model setting (missing values correspond to settings not applicable to the task). All results are from k-fold ($k=5$) evaluations averaged over 10 random runs.}
\label{tab:pred}
\end{table*}

\paragraph{Baselines}  We compare against a simple mean value baseline, as well as against language-wise mean value and model-wise mean value baselines. 
The simple mean value baseline outputs an average of scores $s$ from the training folds for all test entries in the left-out evaluation fold $(i)$ as follows: 
\begin{gather}
    \hat s_{\mathrm{mean}}^{(i)} = \frac{1}{|\mathcal{S} \setminus \mathcal{S}^{(i)}|}  \sum_{s \in \mathcal{S} \setminus \mathcal{S}^{(i)}} s; i \in 1 \dots k
    \label{eq:mean_baseline}
\end{gather}
Note that for tasks involving multiple models, we calculate the RMSE score separately on each model and use the mean RMSE of all models as the final RMSE score.

The language-wise baselines make more informed predictions, taking into account only training instances with the same transfer, source, or target language (depending on the task setting).
For example, the source-language mean value baseline $\hat s_{\mathrm{s\text{-}lang}}^{(i, j)}$ for $j^{\text{th}}$ test instance in fold $i$ outputs an average of the scores $s$ of the training instances that share the \emph{same} source language features $\mathrm{s\text{-}lang}$, as shown in \autoref{eq:language_baseline}: 
\iftrue
\begin{equation}
\begin{gathered}
     \hat s_{\mathrm{s\text{-}lang}}^{(i, j)} = \frac{\sum_{s, \phi} \delta(\phi_{\mathcal{L}, \mathrm{src}}  = \mathrm{s\text{-}lang}) \cdot s}{\sum_{s, \phi} \delta(\phi_{\mathcal{L}, \mathrm{src}} = \mathrm{s\text{-}lang})}   \\
\forall (s, \phi) \in (|\mathcal{S} \setminus \mathcal{S}^{(i)}|, |\Phi \setminus \Phi^{(i)}|) \label{eq:language_baseline}
\end{gathered}
\end{equation}
\fi

\noindent where $\delta$ is the indicator function.
Similarly, we define the target- and the transfer-language mean value baselines.

In a similar manner, we also compare against a model-wise mean value baseline for tasks that include experimental records from multiple models. Now, the prediction for the $j^{\text{th}}$ test instance in the left-out fold $i$ is an average of the scores on the same dataset (as characterized by the language $\phi_{\mathcal{L}}$ and dataset $\phi_{\mathcal{D}}$ features) from all other models:
\iftrue
\begin{equation}
\begin{gathered}
     \hat s_{\mathrm{model}}^{(i, j)} = \frac{\sum_{s, \phi} \delta(\phi_{\mathcal{L}} = \mathrm{lang}, \phi_{\mathcal{D}}=\mathrm{data}) \cdot s}{\sum_{s, \phi} \delta(\phi_{\mathcal{L}} = \mathrm{lang}, \phi_{\mathcal{D}}=\mathrm{data})}   \\
\forall (s, \phi) \in (|\mathcal{S} \setminus \mathcal{S}^{(i)}|, |\Phi \setminus \Phi^{(i)}|) \label{eq:model_baseline}
\end{gathered}
\end{equation}
\fi
\noindent where $\mathrm{lang} = {\Phi^{(i, j)}_{\mathcal{L}}}$ and $\mathrm{data} = \Phi^{(i, j)}_{\mathcal{D}}$ respectively denote the language and dataset features of the test instance.


\paragraph{Main Results}
For multi-model tasks, we can do either \textbf{S}ingle \textbf{M}odel prediction (SM), restricting training and testing of the predictor within a single model, or \textbf{M}ulti-\textbf{M}odel (MM) prediction using a categorical model feature. The RMSE scores of  \textsc{NLPerf} along with the baselines are shown in \autoref{tab:pred}. For all tasks, our single model predictor is able to more accurately estimate the evaluation score of unseen experiments compared to the single model baselines, confirming our hypothesis that the there exists a correlation that can be captured between experimental settings and the downstream performance of NLP systems. The language-wise baselines are much stronger than the simple mean value baseline but still perform worse than our single model predictor. Similarly, the model-wise baseline significantly outperforms the mean value baseline because results from other models reveal much information about the dataset. Even so, our multi-model predictor still outperforms the model-wise baseline.

The results nicely imply that for a wide range of tasks, our predictor is able to reasonably estimate left-out slots in a partly populated table given results of other experiment records, without actually running the system. 

We should note that RMSE scores across different tasks should not be directly compared, mainly because the scale of each evaluation metric is different. For example, a BLEU score~\cite{papineni2002bleu} for evaluating MT experiments typically ranges from~1 to~40, while an accuracy score usually has a much larger range, for example, BLI accuracy ranges from~0.333 to~78.2 and TSF-POS accuracy ranges from~1.84 to~87.98, which consequently makes the RMSE scores of these tasks higher. 

\paragraph{Comparison to Expert Human Performance} 
We constructed a small scale case study to evaluate whether  \textsc{NLPerf} is competitive to the performance of NLP sub-field experts.
We focused on the TED-MT task and recruited 10 MT practitioners,\footnote{None of the study participants were affiliated to the authors' institutions, nor were familiar with this paper's content.} all of whom had published at least~3 MT-related papers in ACL-related conferences.

In the first set of questions, the participants were presented with language pairs from one of the $k$ data folds along with the dataset features and were asked to estimate an eventual BLEU score for each data entry. In the second part of the questionnaire, the participants were tasked with making estimations on the same set of language pairs, but this time they also had access to features, and BLEU scores from all the other folds.\footnote{The interested reader can find an example questionnaire (and make estimations over one of the folds) in the \ref{quiz}.}

 

The partition of the folds is consistent between the human study and the training/evaluation for the predictor. While the first sheet is intended to familiarize the participants with the task, the second sheet fairly adopts the training/evaluation setting for our predictor. As shown in \autoref{tab:human}, our participants outperform the mean baseline even without information from other folds, demonstrating their own strong prior knowledge in the field. In addition, the participants make even more accurate guesses after acquiring more information on experimental records in other folds. In neither case, though, are the human experts competitive to our predictor. In fact, only one of the participants achieved performance comparable to our predictor.
\begin{table}[t]
\centering
\begin{tabular}{lc}
\toprule
Predictor & RMSE \\ \midrule
Mean Baseline & 12.64 \\
Human (w/o training data)   & 9.38 \\
Human (w/ training data)  & 7.29 \\
 \textsc{NLPerf}      & {\bf 6.04} \\ \bottomrule
\end{tabular}
\caption{Our model performs better than human MT experts on the TED-MT prediction task. }
\label{tab:human}
\end{table}

\begin{figure}[t]
  \includegraphics[width=\linewidth]{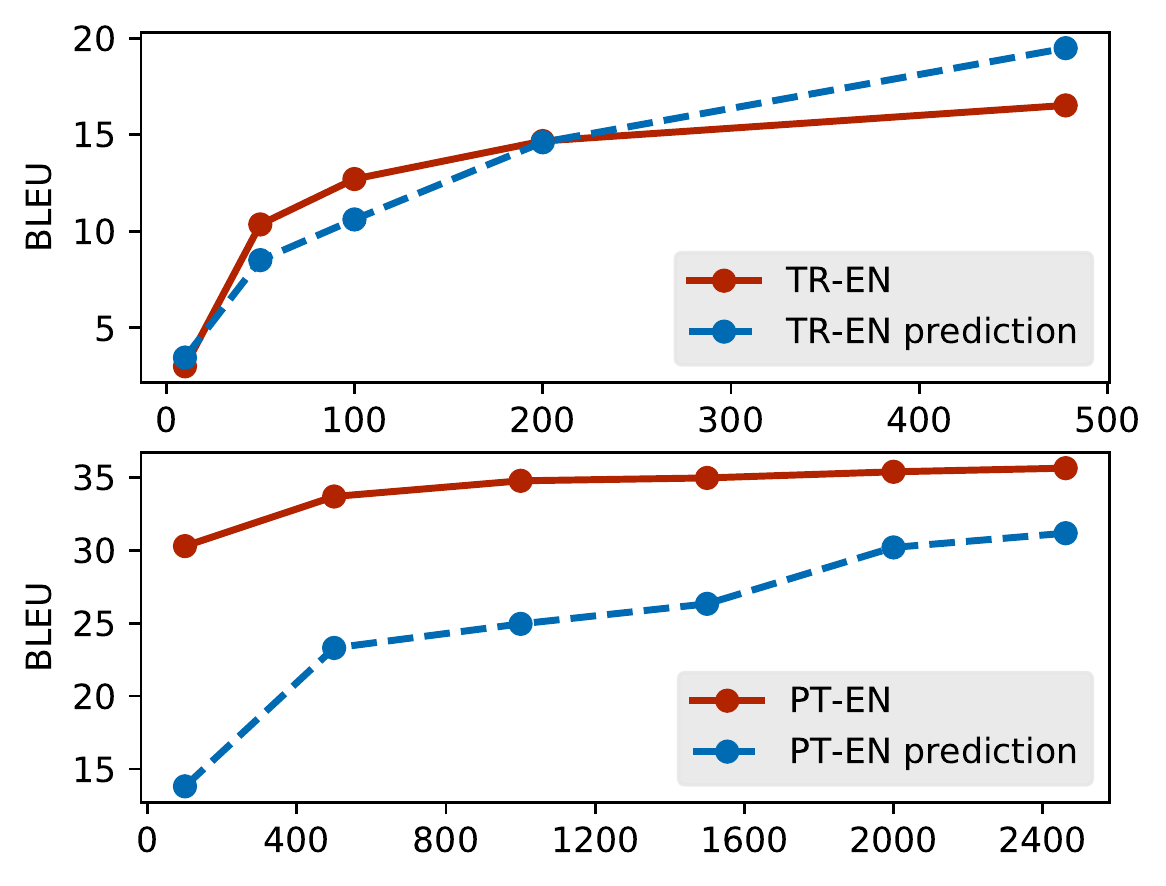}
  \caption{Our model's predicted BLEU scores and true BLEU scores, on sampled \textsc{tr--en} datasets (sizes 10k/50k/100k/200k/478k) and \textsc{pt--en} datasets (sizes 100k/500k/1000k/2000k/2462k), achieving a RMSE score of~1.83 and 9.97 respectively.} 
  \label{fig:f2}
\end{figure}

\begin{figure*}
    \centering
        \includegraphics[width=1\textwidth]{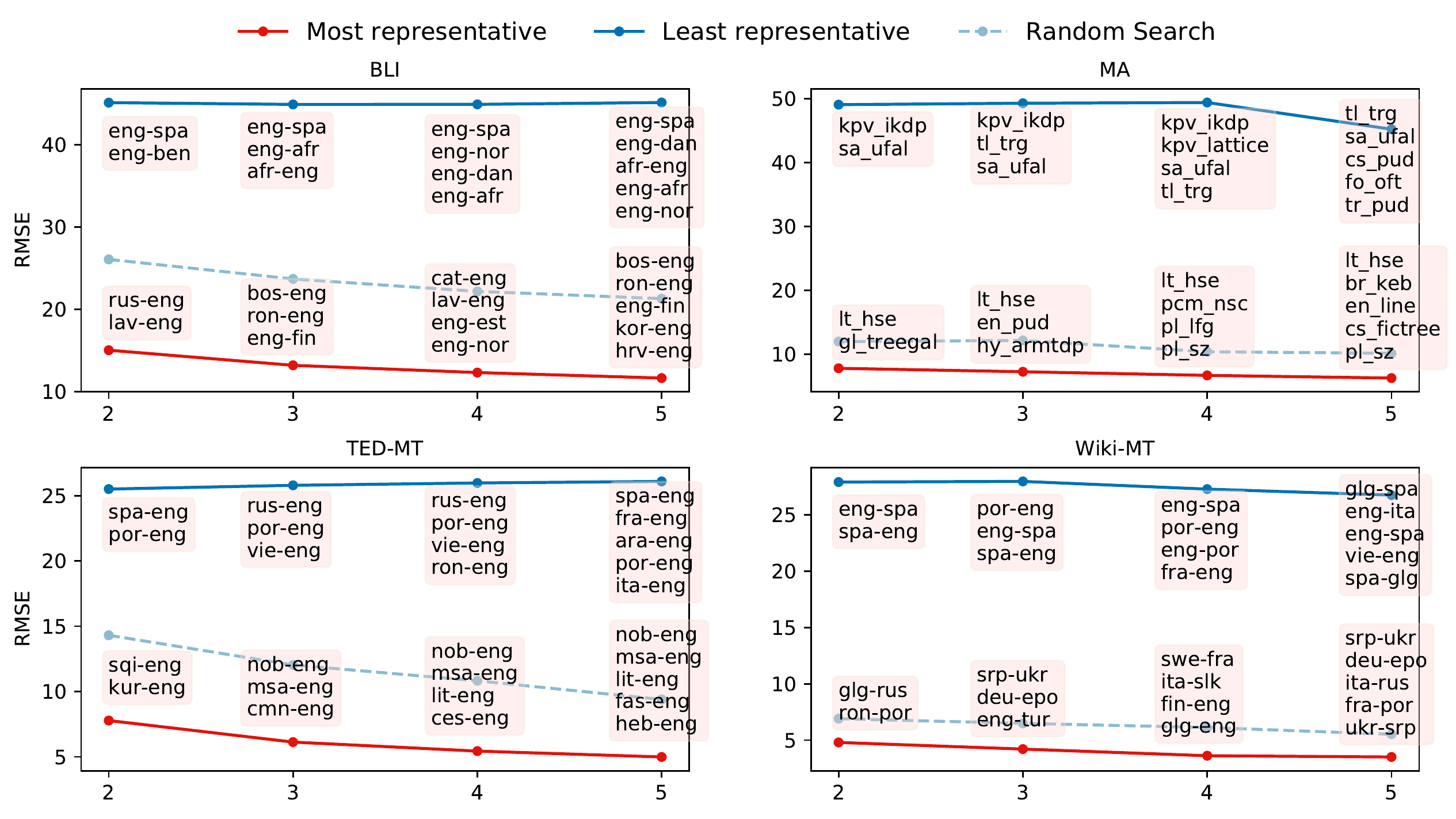}
    \caption{Beam search results (beam size=100) for up to the~5 most (and least) representative datasets for~4 NLP tasks. We also show random search results averaged over 100 random runs.}
    \label{fig:rep}
\end{figure*}

\paragraph{Feature Perturbation} 
Another question of interest concerning predicting performance is ``how will the model perform when trained on data of a different size'' \cite{kolachina-etal-2012-prediction}.
To test  \textsc{NLPerf}'s extrapolation ability in this regard, we conduct an array of experiments on one language pair with various data sizes on the Wiki-MT task. We pick two language pairs, Turkish to English (\textsc{tr--en}) and Portuguese to English (\textsc{pt--en}) as our testbed for the Wiki-MT task. We sample parallel datasets with different sizes and train MT models with each sampled dataset to obtain the true BLEU scores. On the other hand, we collect the features of all sampled datasets and use our predictor (trained over all other languages pairs) to obtain predictions. The plot of true BLEU scores and predicted BLEU scores are shown in Figure~\ref{fig:f2}. Our predictor achieves a very low average RMSE of~1.83 for \textsc{tr--en} pair but a relatively higher RMSE of ~9.97 for \textsc{pt--en} pair. The favorable performance on the tr-en pair demonstrates the possibility of our predictor to do feature extrapolation over data set size. In contrast, the predictions on the pt-en pair are significantly less accurate. This is due to the fact that there are only two other experimental settings scoring as high as 34 BLEU score, with data sizes of 3378k (en-es) and 611k (gl-es), leading to the predictor's inadequacy in predicting high BLEU scores for low-resourced data sets during extrapolation. This reveals the fact that while the predictor is able to extrapolate performance on settings similar to what it has seen in the data, \textsc{NLPerf} may be less successful under circumstances unlike its training inputs.

\section{What Datasets Should We Test On?}
\label{sec:representativeness}

As shown in \autoref{tab:example}, it is common practice to test models on a subset of all available datasets. The reason for this is practical -- it is computationally prohibitive to evaluate on all settings. However, if we pick test sets that are not  \emph{representative} of the data as a whole, we may mistakenly reach unfounded conclusions about how well models perform on other data with distinct properties. For example, models trained on a small-sized dataset may not scale well to a large-sized one, or models that perform well on languages with a particular linguistic characteristic may not do well on languages with other characteristics~\cite{bender2018data}.


Here we ask the following question: if we are only practically able to test on a small number of experimental settings, which ones should we test on to achieve maximally representative results? Answering the question could have practical implications: organizers of large shared tasks like SIGMORPHON \cite{mccarthy-etal-2019-sigmorphon} or UD \cite{zeman-EtAl:2018:K18-2} could create a minimal subset of settings upon which they would ask participants to test to get representative results; similarly, participants could possibly expedite the iteration of model development by testing on the representative subset only. A similar avenue for researchers and companies deploying systems over multiple languages could lead to not only financial savings, but potentially a significant cut-down of emissions from model training~\cite{strubell-etal-2019-energy}. 


We present an approximate explorative solution to the problem mentioned above. Formally, assume that we have a set $\mathcal{N}$, comprising experimental records (both features and scores) of $n$ datasets for one task. We set a number $m$ ($<n$) of datasets that we would like to select as the representative subset. By defining $\mathrm{RMSE}_{\mathcal{A}}(\mathcal{B})$ to be the RMSE score derived from evaluating on one subset $\mathcal{B}$ the predictor trained on another subset of experimental records $\mathcal{A}$, we consider the \emph{most representative} subset $\mathcal{D}$ to be the one that minimizes the RMSE score when predicting all of the other datasets:
\begin{gather}
\operatorname*{arg\,min}_{\mathcal{D} \subset \mathcal{N}}  \mathrm{RMSE}_{\mathcal{D}}(\mathcal{N}\setminus {\mathcal{D}}) .
\label{eq:min}
\end{gather}


Naturally, enumerating all $n \choose m$ possible subsets would be prohibitively costly, even though it would lead to the optimal solution. Instead, we employ a beam-search-like approach to efficiently search for an approximate solution to the best performing subset of arbitrary size. Concretely, we start our approximate search with an exhaustive enumeration of all subsets of size~2. At each following step $t$, we only consider the best $k$ subsets $\{\mathcal{D}_t^{(i)} ; i \in 1, \dots, k\}$ into account and discard the rest. As shown in \autoref{eq:expand}, for each candidate subset, we expand it with one more data point,
\begin{gather}
     \{\mathcal{D}^{{(i)}}_t \cup \{s\}; \forall i \in 1 \dots k, s \in \mathcal{N} \setminus \mathcal{D}^{(i)}_t\}.
  \label{eq:expand}
\end{gather}

For tasks that involve multiple models,  we take experimental records of the selected dataset from all models into account during expansion. Given all expanded subsets, we train a predictor for each to evaluate on the rest of the data sets, and keep the best performing $k$ subsets $\{\mathcal{D}^{(i)}_{t+1}; i \in 1, \dots, k\}$ with minimum RMSE scores for the next step.   Furthermore, note that by simply changing the $\arg\min$ to an $\arg\max$ in \autoref{eq:min}, we can also find the \textit{least} representative datasets.

We present search results for four tasks\footnote{Readers can find results on other tasks in \autoref{appendix:rep}.} as beam search progresses in Figure~\ref{fig:rep}, with corresponding RMSE scores from all remaining datasets as the y-axis. For comparison, we also conduct random searches by expanding the subset with a randomly selected experimental record. In all cases, the most representative sets are an aggregation of datasets with diverse characteristics such as languages and dataset sizes. For example, in the Wiki-MT task, the 5 most representative datasets include languages that fall into a diverse range of language families such as Romance, Turkic, Slavic, etc. while the least representative ones include duplicate pairs (opposite directions) mostly involving English. The phenomenon is more pronounced in the TED-MT task, where not only the 5 most representative source languages are diverse, but also the dataset sizes. Specifically, the Malay-English (msa-eng) is a tiny dataset (5k parallel sentences), and Hebrew-English (heb-eng) is a high-resource case (212k parallel sentences). 

Notably, for BLI task, to test how representative the commonly used datasets are, we select the most frequent 5 language pairs shown in \autoref{tab:example}, namely en-de, es-en, en-es, fr-en, en-fr for evaluation. Unsurprisingly, we get an RMSE score as high as $43.44$, quite close to the performance of the worst representative set found using beam search. This finding indicates that the standard practice of choosing datasets for evaluation is likely unrepresentative of results over the full dataset spectrum, well aligned with the claims in \citet{anastasopoulos20embeddings}.

A particularly encouraging observation is that the predictor trained with only the 5 most representative datasets can achieve an RMSE score comparable to k-fold validation, which required using all of the datasets for training.\footnote{to be accurate, $k-1$ folds of all datasets.}  This indicates that one would only need to train NLP models on a small set of \textit{representative} datasets to obtain reasonably plausible predictions for the rest.

\section{Can We Extrapolate Performance for New Models?}
\label{sec:model}

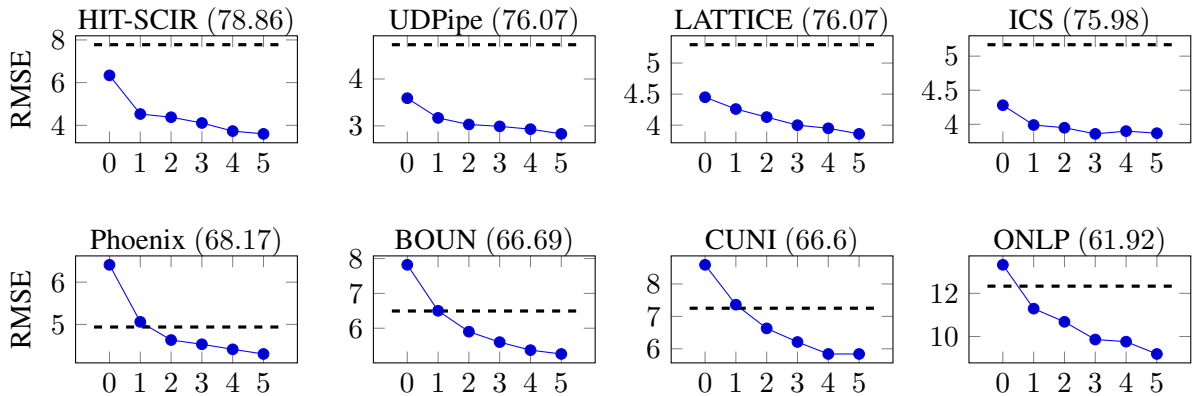
\begin{figure*}[t]
\centering
\pgfplotstableread[row sep=\\,col sep=&]{RMSE & HIT-SCIR&UDPipe&LATTICE&ICS&Phoenix&BOUN&CUNI&ONLP\\0 & 17.848847975246006&17.418067963717686&16.28693224406152&18.227522242344616&17.740209920074914&18.86462308635162&16.311427688913433&18.945531088275875\\
1 & 18.137504742919024&17.51446571791011&16.592203920326742&18.075620079961336&16.865106735738873&19.174635869593594&15.767351508710806&18.500509900688357\\
2 & 17.554779021150626&17.47139395230915&16.19438960654364&17.464743393737432&17.588569988109807&18.614915002962764&16.43884829117103&19.302245690439666\\
3 & 17.86384708160511&17.284911438808443&16.50147517818397&18.406994286568718&16.850382282065457&18.548214427265503&16.89685688627162&19.67517066584187\\
4 & 17.61188335382808&17.301604398796158&16.22851358840078&18.03742608265518&17.611166096388526&18.5587580291055&16.38793434039233&19.47595932122629\\
5 & 18.050420604994848&17.38200618360415&16.413253857488897&18.172512646918808&17.7351604398222&18.025383810850553&16.60846694185999&19.81920190347248\\
}\basedata
\pgfplotstableread[row sep=\\,col sep=&]{RMSE & HIT-SCIR&UDPipe&LATTICE&ICS&Phoenix&BOUN&CUNI&ONLP\\0 & 6.34&3.59&4.45&4.28&6.41&7.82&8.59&13.32\\
1 & 4.53&3.17&4.26&3.99&5.06&6.5&7.36&11.29\\
2 & 4.38&3.03&4.13&3.95&4.63&5.9&6.63&10.68\\
3 & 4.11&2.99&4.0&3.86&4.53&5.6&6.21&9.86\\
4 & 3.73&2.93&3.95&3.9&4.41&5.37&5.84&9.76\\
5 & 3.6&2.83&3.86&3.87&4.3&5.26&5.84&9.19\\
}\systemdata
\pgfplotstableread[row sep=\\,col sep=&]{RMSE & HIT-SCIR&UDPipe&LATTICE&ICS&Phoenix&BOUN&CUNI&ONLP\\
-0.5 &7.77594767584605&4.730330712224502&5.295482850915353&5.167684860538444&4.9381156775584305&6.493538408932312&7.25086131320416&12.331740524509089\\
5.5 &7.77594767584605&4.730330712224502&5.295482850915353&5.167684860538444&4.9381156775584305&6.493538408932312&7.25086131320416&12.331740524509089\\
}\basedataa
\begin{tikzpicture}
\begin{groupplot}[group style={group size=4 by 2,vertical sep=1.5cm},height=3cm,width=4.5cm,title style={yshift=-2ex,}]
\nextgroupplot[title=HIT-SCIR $(78.86)$,xtick=data,ylabel=RMSE]
\addplot+ table[x=RMSE,y=HIT-SCIR]{\systemdata};
\addplot[mark=none, black, very thick, dashed] table[x=RMSE,y=HIT-SCIR]{\basedataa};
\nextgroupplot[title=UDPipe $(76.07)$,xtick=data]
\addplot+ table[x=RMSE,y=UDPipe]{\systemdata};
\addplot[mark=none, black, very thick, dashed] table[x=RMSE,y=UDPipe]{\basedataa};
\nextgroupplot[title=LATTICE $(76.07)$,xtick=data]
\addplot+ table[x=RMSE,y=LATTICE]{\systemdata};
\addplot[mark=none, black, very thick, dashed] table[x=RMSE,y=LATTICE]{\basedataa};
\nextgroupplot[title=ICS $(75.98)$,xtick=data]
\addplot+ table[x=RMSE,y=ICS]{\systemdata};
\addplot[mark=none, black, very thick, dashed] table[x=RMSE,y=ICS]{\basedataa};
\nextgroupplot[title=Phoenix $(68.17)$,xtick=data,ylabel=RMSE]
\addplot+ table[x=RMSE,y=Phoenix]{\systemdata};
\addplot[mark=none, black, very thick, dashed] table[x=RMSE,y=Phoenix]{\basedataa};
\nextgroupplot[title=BOUN $(66.69)$,xtick=data]
\addplot+ table[x=RMSE,y=BOUN]{\systemdata};
\addplot[mark=none, black, very thick, dashed] table[x=RMSE,y=BOUN]{\basedataa};
\nextgroupplot[title=CUNI $(66.6)$,xtick=data]
\addplot+ table[x=RMSE,y=CUNI]{\systemdata};
\addplot[mark=none, black, very thick, dashed] table[x=RMSE,y=CUNI]{\basedataa};
\nextgroupplot[title=ONLP $(61.92)$,xtick=data]
\addplot+ table[x=RMSE,y=ONLP]{\systemdata};
\addplot[mark=none, black, very thick, dashed] table[x=RMSE,y=ONLP]{\basedataa};
\end{groupplot}
\end{tikzpicture}
\caption{RMSE scores of \textbf{UD} task from dataset-wise mean value predictor (the dashed black line in each graph) and predictors trained with experimental records of other models and~0--5 records from a new model.}
\label{fig:UD_label}
\end{figure*}
In another common scenario, researchers propose new models for an existing task. It is both time-consuming and computationally intensive to run experiments with all settings for a new model. In this section, we explore if we can use past experimental records from other models and a minimal set of experiments from the new model to give a plausible prediction over the rest of the datasets, potentially reducing the time and resources needed for experimenting with the new model to a large extent. We use the task of UD parsing as our testbed\footnote{MA and BLI task results are in \autoref{app:newmodel}} as it is the task with most unique models (25 to be exact). Note that we still only use a single categorical feature for the model type. 

To investigate how many experiments are needed to have a plausible prediction for a new model, we first split the experimental records equally into a sample set and a test set. Then we randomly sample $n \ (0 \leq n \leq 5)$ experimental records from the sample set and add them into the collection of experiment records of past models. Each time we re-train a predictor and evaluate on the test set. The random split repeats 50 times and the random sampling repeats 50 times, adding up to a total of 2500 experiments. We use the mean value of the results from other models, shown in \autoref{eq:mean} as the prediction baseline for the left-out model, and because experiment results of other models reveal significant information about the dataset, this serves as a relatively strong baseline:
\begin{gather}
    \hat{s}_k = \frac{1}{n-1} \sum_{i=1}^n \mathbbm{1} (i \in \mathcal{M} / \{k\}) \cdot s_i.
    \label{eq:mean}
\end{gather}
\noindent $\mathcal{M}$ denotes a collection of models and $k$ denotes the left-out model.

We show the prediction performance (in RMSE) over 8 systems\footnote{The best and worst~4 systems from the shared task.} in \autoref{fig:UD_label}. Interestingly, the predictor trained with no model records (0) outperforms the mean value baseline for the~4 best systems, while it is the opposite case on the 4 worst systems. Since there is no information provided about the new-coming model, the predictions are solely based on dataset and language features. One reason might explain the phenomenon - the correlation between the features and the scores of the worse-performing systems is different from those better-performing systems, so the predictor is unable to generalize well (ONLP).


In the following discussion, we use RMSE@n to denote the RMSE from the predictor trained with $n$ data points of a new model. The relatively low RMSE@0 scores indicate that other models' features and scores are informative for predicting the performance of the new model even without new model information. Comparing RMSE@0 and RMSE@1, we observe a consistent improvement for almost all systems, indicating that \textsc{NLPerf} trained on even a single extra random example achieves more accurate estimates over the test sets. Adding more data points consistently leads to additional gains. However, predictions on worse-performing systems benefit more from it than for better-performing systems, indicating that their feature-performance correlation might be considerably different. The findings here indicate that by extrapolating from past experiments, one can make plausible judgments for newly developed models.


\section{Related Work}
\label{sec:related}
As discusssed in \citet{domhan2015speeding}, there are two main threads of work focusing on predicting performance of machine learning algorithms. The first thread is to predict the performance of a method as a function of its training time, while the second thread is to predict a method's performance as a function of the training dataset size. Our work belongs in the second thread, but could easily be extended to encompass training time/procedure.

In the first thread, \citet{kolachina2012prediction} attempt to infer learning curves based on training data features and extrapolate the initial learning curves based on BLEU measurements for statistical machine translation (SMT). By extrapolating the performance of initial learning curves, the predictions on the remainder allows for early termination of a bad run~\cite{domhan2015speeding}.

In the second thread, \citet{birch2008predicting} adopt linear regression to capture the relationship between data features and SMT performance and find that the amount of reordering, the morphological
complexity of the target language and the relatedness of the two languages explains the majority of performance variability. More recently, \citet{elsahar2019annotate} use domain shift metrics such as $\mathcal{H}$-divergence based metrics to predict drop in performance under domain-shift. \citet{rosenfeld2020a} explore the functional form of the dependency of the generalization error of neural models on model and data size. We view our work as a generalization of such approaches, appropriate for application on any NLP task.

\section{Conclusion and Future Work}
\label{sec:conclusion}
In this work, we investigate whether the experiment setting itself is informative for predicting the evaluation scores of NLP tasks. Our findings promisingly show that given a sufficient number of past training experimental records, our predictor can 1) outperform human experts; 2) make plausible predictions even over new-coming models and languages; 3) extrapolate well on features like dataset size; 4) provide a guide on how we should choose representative datasets for fast iteration. 

While this discovery is a promising start, there are still several avenues on improvement in future work.

First, the dataset and language settings covered in our study are still limited. Experimental records we use are from relatively homogeneous settings, e.g. all datasets in Wiki-MT task are sentencepieced to have 5000 subwords, indicating that our predictor may fail for other subword settings. Our model also failed to generalize to cases where feature values are out of the range of the training experimental records. We attempted to apply the predictor of Wiki-MT to evaluate on a low-resource MT dataset, translating from Mapudungun (arn) to Spanish (spa) with the dataset from~\citet{duan19resource}, but ended up with a poor RMSE score. It turned out that the average sentence length of the arn--spa data set is much lower than that of the training data sets and our predictors fail to generalize to this different setting. 

Second, using a categorical feature to denote model types constrains its expressive power for modeling performance. In reality, a slight change in model hyperparameters \cite{hoos2014efficient, probst2019tunability}, optimization algorithms \cite{kingma2014adam}, or even random seeds~\cite{madhyastha-jain-2019-model} may give rise to a significant variation in performance, which our predictor is not able to capture. While investigating the systematic implications of model structures or hyperparameters is practically infeasible in this study, we may use additional information such as textual model descriptions for modeling NLP models and training procedures more elaborately in the future.

Lastly, we assume that the distribution of training and testing data is the same, which does not consider domain shift. On top of this, there might also be a domain shift between data sets of training and testing experimental records. We believe that modeling domain shift is a promising future direction to improve performance prediction.

\section*{Acknowledgement}
\label{sec:acknowledgement}
The authors sincerely thank all the reviewers for their insightful comments and suggestions, Philipp Koehn, Kevin Duh, Matt Post, Shuoyang Ding, Xuan Zhang, Adi Renduchintala, Paul McNamee, Toan Nguyen and Kenton Murray for conducting human evaluation for the TED-MT task, Daniel Beck for discussions on Gaussian Processes, Shruti Rijhwani, Xinyi Wang, Paul Michel for discussions on this paper. This work is generously supported from the National Science Foundation under grant 1761548.

\bibliography{acl2020}
\bibliographystyle{acl_natbib}

\clearpage
\newpage
\appendix
\section*{Appendix}
\section{Questionnaire}
\label{quiz}
\newcolumntype{a}{>{\columncolor{yellow}}c}

An example of the first questionnaire from our user case study is shown below. The second sheet also included the results in~44 more language pairs. 
We provide an answer key after the second sheet.

\begin{table}[h]
\begin{tabular}{lllccccca}
\toprule
\multicolumn{9}{l}{Please provide your prediction of the BLEU score based on the language pair and features of the dataset } \\ 
\multicolumn{9}{l}{(the domain of the training and test sets is TED talks). After you finish, please go to sheet v2.} \\ \midrule
idx  & Source  & Target  & Parallel & Source   & Source    & Target & Target  & BLEU \\ 
& Language & Language  & Sentences & vocab  & subword  & vocab  & subword & \\
 &  &  & (k) &size (k)  & vocab  & size (k)  &  vocab  & \\
  &  &  & &  &  size (k) & &   size( k) & \\

\midrule
1                                                                                                                                                 & Basque (eus)          & English                   & 5                      & 20                         & 8                                  & 9                          & 6                                  &     \\
2                                                                                                                                                 & Slovak (slk)         & English                  & 61                     & 134                        & 8                                  & 36                         & 8                                  &     \\
3                                                                                                                                                 & Burmese   (mya)      & English                    & 21                     & 101                        & 8                                  & 21                         & 8                                  &  \\
4                                                                                                                                                 & Korean   (kor)       & English                         & 206                    & 386                        & 9                                  & 67                         & 8                                  &      \\
5                                                                                                                                                 & Lithuanian  (lit)    & English                     & 42                     & 108                        & 8                                  & 29                         & 8                                  &      \\
6                                                                                                                                                 & Arabic   (ara)       & English                      & 214                    & 308                        & 8                                  & 69                         & 8                                  &      \\
7                                                                                                                                                 & Czech (ces)          & English                      & 103                    & 181                        & 8                                  & 47                         & 8                                  &      \\
8                                                                                                                                                 & Esperanto   (epo)    & English                     & 7                      & 21                         & 8                                  & 10                         & 6                                  &      \\
9                                                                                                                                                 & Finnish (fin)        & English                     & 24                     & 77                         & 8                                  & 22                         & 8                                  &      \\
10                                                                                                                                                & Albanian   (sqi)     & English               & 45                     & 93                         & 8                                  & 30                         & 8                                  &      \\
11                                                                                                                                                & Vietnamese (vie)     & English                & 172                    & 66                         & 8                                  & 61                         & 8                                  &      \\ \bottomrule
\end{tabular}
\end{table}
\clearpage
\newpage

\begin{table}[h]
\begin{tabular}{lllccccca}
\toprule
\multicolumn{9}{l}{Please provide your prediction of the BLEU score in the yellow area given all the information in this  } \\ 
\multicolumn{9}{l}{sheet. Note that all experiments are trained with the same model.} \\ \midrule
idx  & Source  & Target  & Parallel & Source   & Source    & Target & Target  & BLEU \\ 
& Language & Language  & Sentences & vocab  & subword  & vocab  & subword & \\
&  &  & (k) &size (k)  & vocab  & size (k)  &  vocab  & \\
 &  &  & &  &  size (k) & &   size( k) & \\ \midrule
1   & Basque (eus)           & English         & 5                      & 20                         & 8                                  & 9                          & 6                                  &       \\
2   & Slovak (slk)           & English         & 61                     & 134                        & 8                                  & 36                         & 8                                  &       \\
3   & Burmese (mya)          & English         & 21                     & 101                        & 8                                  & 21                         & 8                                  &       \\
4   & Korean (kor)           & English         & 206                    & 386                        & 9                                  & 67                         & 8                                  &       \\
5   & Lithuanian (lit)       & English         & 42                     & 108                        & 8                                  & 29                         & 8                                  &       \\
6   & Arabic (ara)           & English         & 214                    & 308                        & 8                                  & 69                         & 8                                  &       \\
7   & Czech (ces)            & English         & 103                    & 181                        & 8                                  & 47                         & 8                                  &       \\
8   & Esperanto (epo)        & English         & 7                      & 21                         & 8                                  & 10                         & 6                                  &       \\
9   & Finnish (fin)          & English         & 24                     & 77                         & 8                                  & 22                         & 8                                  &       \\
10  & Albanian (sqi)         & English         & 45                     & 93                         & 8                                  & 30                         & 8                                  &       \\
11  & Vietnamese (vie)       & English         & 172                    & 66                         & 8                                  & 61                         & 8                                  &       \\
12  & French (fra)           & English         & 192                    & 158                        & 8                                  & 65                         & 8                                  & 37.74 \\
13  & Estonian (est)         & English         & 11                     & 39                         & 8                                  & 14                         & 7                                  & 9.9   \\
14  & Macedonian (mkd)       & English         & 25                     & 61                         & 8                                  & 23                         & 8                                  & 21.75 \\
15  & Bosnian (bos)          & English         & 6                      & 23                         & 8                                  & 9                          & 6                                  & 32.42 \\
16  & Swedish (swe)          & English         & 57                     & 84                         & 8                                  & 34                         & 8                                  & 33.92 \\
17  & Polish (pol)           & English         & 176                    & 267                        & 8                                  & 63                         & 8                                  & 21.51 \\
18  & Persian (fas)          & English         & 151                    & 148                        & 8                                  & 57                         & 8                                  & 24.5  \\
19  & Kurdish (kur)          & English         & 10                     & 39                         & 8                                  & 14                         & 7                                  & 6.86  \\
20  & Hungarian (hun)        & English         & 147                    & 305                        & 8                                  & 56                         & 8                                  & 22.67 \\
21  & Slovenian (slv)        & English         & 20                     & 58                         & 8                                  & 20                         & 8                                  & 14.18 \\
22  & Romanian (ron)         & English         & 181                    & 205                        & 8                                  & 63                         & 8                                  & 32.42 \\
23  & Russian (rus)          & English         & 208                    & 291                        & 8                                  & 68                         & 8                                  & 22.6  \\
24  & Serbian (srp)          & English         & 137                    & 239                        & 8                                  & 54                         & 8                                  & 30.41 \\
25  & Tamil (tam)            & English         & 6                      & 27                         & 8                                  & 10                         & 6                                  & 1.82  \\
26  & Kazakh (kaz)           & English         & 3                      & 15                         & 8                                  & 7                          & 5                                  & 2.05  \\
27  & Marathi (mar)          & English         & 10                     & 29                         & 8                                  & 13                         & 7                                  & 3.68  \\
28  & Ukrainian (ukr)        & English         & 108                    & 191                        & 8                                  & 48                         & 8                                  & 24.09 \\
29  & Thai (tha)             & English         & 98                     & 323                        & 8                                  & 45                         & 8                                  & 20.34 \\
30  & Belarusian (bel)       & English         & 5                      & 20                         & 8                                  & 8                          & 5                                  & 2.85  \\
31  & Turkish (tur)          & English         & 182                    & 304                        & 8                                  & 63                         & 8                                  & 22.52 \\
32  & Azerbaijani (aze)      & English         & 6                      & 23                         & 8                                  & 9                          & 6                                  & 3.1   \\
33  & German (deu)           & English         & 168                    & 194                        & 8                                  & 61                         & 8                                  & 33.15 \\
34  & Bulgarian (bul)        & English         & 174                    & 216                        & 8                                  & 62                         & 8                                  & 35.78 \\
35  & Norwegian (nob) & English         & 16                     & 36                         & 8                                  & 17                         & 7                                  & 29.63 \\
36  & Georgian (kat)         & English         & 13                     & 44                         & 8                                  & 15                         & 7                                  & 4.94  \\
37  & Danish (dan)           & English         & 45                     & 72                         & 8                                  & 31                         & 8                                  & 37.73 \\
38  & Armenian (hye)         & English         & 21                     & 56                         & 8                                  & 20                         & 8                                  & 13.97 \\
39  & Mandarin (cmn)         & English         & 200                    & 481                        & 9                                  & 67                         & 8                                  & 17.0  \\ \bottomrule

\end{tabular}
\end{table}
\clearpage
\newpage
\begin{table}[h]
\begin{tabular}{lllccccca}
\toprule
idx  & Source  & Target  & Parallel & Source   & Source    & Target & Target  & BLEU \\ 
& Language & Language  & Sentences & vocab  & subword  & vocab  & subword & \\ \midrule
40  & Indonesian (ind)       & English         & 87                     & 76                         & 8                                  & 43                         & 8                                  & 27.27 \\
41  & Galician (glg)         & English         & 10                     & 28                         & 8                                  & 13                         & 7                                  & 16.84 \\
42  & Portuguese (por)       & English         & 185                    & 165                        & 8                                  & 64                         & 8                                  & 41.67 \\
43  & Urdu (urd)             & English         & 6                      & 13                         & 6                                  & 10                         & 6                                  & 3.38  \\
44  & Italian (ita)          & English         & 205                    & 195                        & 8                                  & 67                         & 8                                  & 35.67 \\
45  & Spanish (spa)          & English         & 196                    & 179                        & 8                                  & 66                         & 8                                  & 39.48 \\
46  & Greek (ell)            & English         & 134                    & 171                        & 8                                  & 54                         & 8                                  & 34.94 \\
47  & Bengali (ben)          & English         & 5                      & 18                         & 8                                  & 9                          & 6                                  & 2.79  \\
48  & Japanese (jpn)         & English         & 204                    & 584                        & 9                                  & 67                         & 8                                  & 11.42 \\
49  & Malay (msa)            & English         & 5                      & 13                         & 7                                  & 9                          & 6                                  & 3.68  \\
50  & Dutch (nld)            & English         & 184                    & 172                        & 8                                  & 63                         & 8                                  & 34.27 \\
51  & Croatian (hrv)         & English         & 122                    & 191                        & 8                                  & 52                         & 8                                  & 31.84 \\
52  & Hebrew (heb)           & English         & 212                    & 276                        & 8                                  & 68                         & 8                                  & 33.89 \\
53  & Mongolian (mon)        & English         & 8                      & 21                         & 8                                  & 11                         & 6                                  & 2.96  \\
54  & Hindi (hin)            & English         & 19                     & 31                         & 8                                  & 19                         & 7                                  & 14.25 \\ \bottomrule
\end{tabular}
\end{table}

\noindent
\rotatebox[origin=c]{180}{%
\noindent
\begin{minipage}[h]{2\linewidth}
Answer Key: eus: 3.37, slk: 25.36, mya: 3.93, kor: 16.23, lit: 13.75, ara: 28.38, ces: 25.07, epo: 3.28, fin: 13.79, sqi: 29.6, vie: 24.67.
\end{minipage}%
}%
\clearpage

\onecolumn
\section{Representative datasets}
\label{appendix:rep}
In this section, we show the searching results of most/least representative subsets for the rest of the five tasks.

\begin{figure*}[h!]
    \centering
    \includegraphics[width=0.97\textwidth]{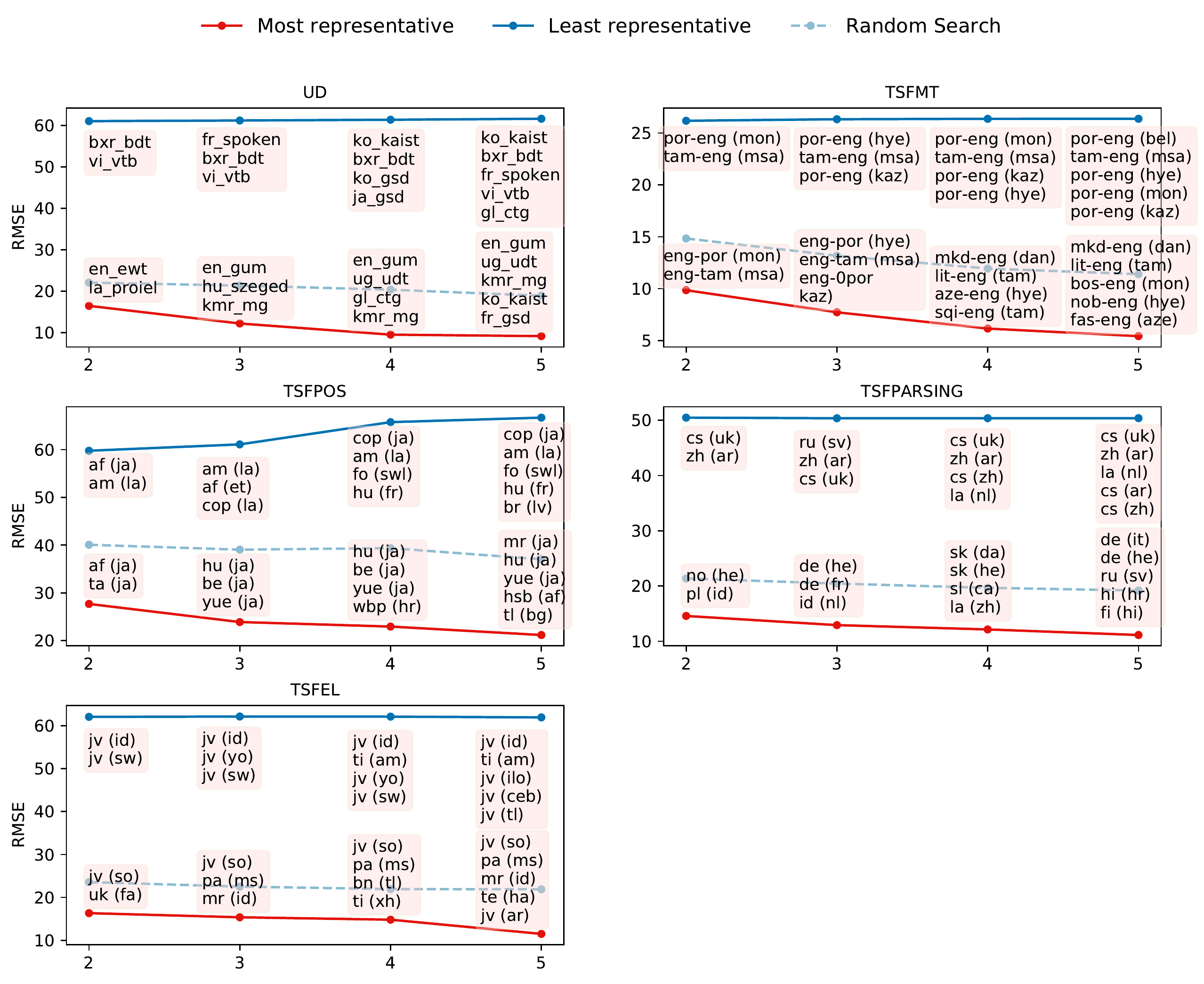}
    \caption {Beam search results (beam size=100) for up to the~5 most (and least) representative datasets for the remaining NLP tasks. We also show random search results of corresponding sizes.}
    \label{fig:rep2}
\end{figure*}

\clearpage
\section{New Model}
\label{app:newmodel}

In this section, we show the extrapolation performance for new models on BLI, MA and the remaining systems of UD.

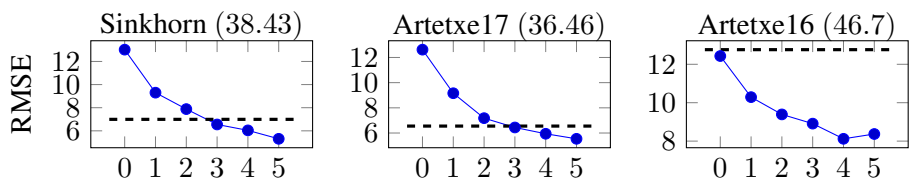
\begin{figure*}[h!]
\centering
\pgfplotstableread[row sep=\\,col sep=&]{RMSE & Sinkhorn&Artetxe17&Artetxe16\\0 & 22.179447354643056&21.618807728588326&18.555778393649334\\
1 & 21.68458917306349&21.46464305388181&18.218779493263067\\
2 & 22.04300312413625&21.124495781365383&18.367238619586086\\
3 & 21.79235122161272&21.467969436153073&18.407385167488467\\
4 & 21.9803597495183&21.489170870682113&18.74314730895624\\
5 & 21.686453761520177&21.921701387246042&18.606382354226966\\
}\basedata
\pgfplotstableread[row sep=\\,col sep=&]{RMSE & Sinkhorn&Artetxe17&Artetxe16\\0 & 13.02&12.62&12.43\\
1 & 9.3&9.16&10.29\\
2 & 7.88&7.18&9.39\\
3 & 6.55&6.44&8.91\\
4 & 6.04&5.94&8.12\\
5 & 5.31&5.54&8.37\\
}\systemdata
\pgfplotstableread[row sep=\\,col sep=&]{RMSE & Sinkhorn&Artetxe17&Artetxe16\\
-0.5 &6.9930963650123426&6.550839631077304&12.761657494342456\\
5.5 &6.9930963650123426&6.550839631077304&12.761657494342456\\
}\basedataa
\begin{tikzpicture}
\begin{groupplot}[group style={group size=3 by 1,vertical sep=1.5cm},height=3cm,width=4.5cm,title style={yshift=-2ex,}]
\nextgroupplot[title=Sinkhorn $(38.43)$,xtick=data,ylabel=RMSE]
\addplot+ table[x=RMSE,y=Sinkhorn]{\systemdata};
\addplot[mark=none, black, very thick, dashed] table[x=RMSE,y=Sinkhorn]{\basedataa};
\nextgroupplot[title=Artetxe17 $(36.46)$,xtick=data]
\addplot+ table[x=RMSE,y=Artetxe17]{\systemdata};
\addplot[mark=none, black, very thick, dashed] table[x=RMSE,y=Artetxe17]{\basedataa};
\nextgroupplot[title=Artetxe16 $(46.7)$,xtick=data]
\addplot+ table[x=RMSE,y=Artetxe16]{\systemdata};
\addplot[mark=none, black, very thick, dashed] table[x=RMSE,y=Artetxe16]{\basedataa};
\end{groupplot}
\end{tikzpicture}
\caption{RMSE scores of \textbf{BLI} task from dataset-wise mean value predictor (the dashed black line in each graph) and predictors trained with experimental records of other models and~0--5 records from a new model (as indicated by the title of each graph).}
\label{fig:BLI_label}
\end{figure*}

\begin{figure*}[h!]
\centering
\pgfplotstableread[row sep=\\,col sep=&]{RMSE & CHARLES-SAARLAND-02-2&Unknown&EDINBURGH-01-2&OHIOSTATE-01-2&CMU-01-2-DataAug&CARNEGIEMELLON-02-2\\0 & 8.806207336308526&8.258653974007615&8.011381827725355&10.243627945156595&9.988686736070077&14.187222124823629\\
1 & 8.543097351763901&8.16481437645381&8.555316760494797&10.505926132440372&10.176454455183435&14.37756534835272\\
2 & 8.950726713236799&8.328195443083617&8.536643614649575&10.524461654816573&10.3074265115832&13.23965568042148\\
3 & 8.803501852312877&8.199334118546611&7.838438493272847&10.584731968223673&10.264389353948502&13.530426726422089\\
4 & 8.446846492554995&8.527958778101478&8.104147204560965&10.341511729604294&9.892910516823017&13.372742436902309\\
5 & 8.693806560387527&8.18664875573306&8.190298893933749&9.866451167896708&10.271353265509088&13.711405337526777\\
}\basedata
\pgfplotstableread[row sep=\\,col sep=&]{RMSE & CHARLES-SAARLAND-02-2&Unknown&EDINBURGH-01-2&OHIOSTATE-01-2&CMU-01-2-DataAug&CARNEGIEMELLON-02-2\\0 & 6.1&6.39&1.98&5.09&7.01&8.53\\
1 & 4.44&4.83&2.11&4.54&6.08&8.41\\
2 & 4.03&4.46&2.16&4.57&6.18&7.28\\
3 & 3.75&4.26&2.12&4.33&5.94&7.1\\
4 & 3.62&3.95&2.16&4.39&5.47&6.48\\
5 & 3.5&3.7&2.14&4.01&5.55&6.48\\
}\systemdata
\pgfplotstableread[row sep=\\,col sep=&]{RMSE & CHARLES-SAARLAND-02-2&Unknown&EDINBURGH-01-2&OHIOSTATE-01-2&CMU-01-2-DataAug&CARNEGIEMELLON-02-2\\
-0.5 &6.03957435833905&6.3767533971332515&2.3908937757100333&3.5008121621121764&5.570983580668209&7.467776646190161\\
5.5 &6.03957435833905&6.3767533971332515&2.3908937757100333&3.5008121621121764&5.570983580668209&7.467776646190161\\
}\basedataa
\begin{tikzpicture}
\begin{groupplot}[group style={group size=3 by 2,vertical sep=1.5cm, horizontal sep=2cm },height=3cm,width=4.5cm,title style={yshift=-2ex,}]
\nextgroupplot[title=CHARLES-SAARLAND-02-2 $(93.23)$,xtick=data,ylabel=RMSE]
\addplot+ table[x=RMSE,y=CHARLES-SAARLAND-02-2]{\systemdata};
\addplot[mark=none, black, very thick, dashed] table[x=RMSE,y=CHARLES-SAARLAND-02-2]{\basedataa};
\nextgroupplot[title=Unknown $(93.19)$,xtick=data]
\addplot+ table[x=RMSE,y=Unknown]{\systemdata};
\addplot[mark=none, black, very thick, dashed] table[x=RMSE,y=Unknown]{\basedataa};
\nextgroupplot[title=EDINBURGH-01-2 $(88.93)$,xtick=data]
\addplot+ table[x=RMSE,y=EDINBURGH-01-2]{\systemdata};
\addplot[mark=none, black, very thick, dashed] table[x=RMSE,y=EDINBURGH-01-2]{\basedataa};
\nextgroupplot[title=OHIOSTATE-01-2 $(87.42)$,xtick=data,ylabel=RMSE]
\addplot+ table[x=RMSE,y=OHIOSTATE-01-2]{\systemdata};
\addplot[mark=none, black, very thick, dashed] table[x=RMSE,y=OHIOSTATE-01-2]{\basedataa};
\nextgroupplot[title=CMU-01-2-DataAug $(86.53)$,xtick=data]
\addplot+ table[x=RMSE,y=CMU-01-2-DataAug]{\systemdata};
\addplot[mark=none, black, very thick, dashed] table[x=RMSE,y=CMU-01-2-DataAug]{\basedataa};
\nextgroupplot[title=CARNEGIEMELLON-02-2 $(85.06)$,xtick=data]
\addplot+ table[x=RMSE,y=CARNEGIEMELLON-02-2]{\systemdata};
\addplot[mark=none, black, very thick, dashed] table[x=RMSE,y=CARNEGIEMELLON-02-2]{\basedataa};
\end{groupplot}
\end{tikzpicture}
\caption{RMSE scores of \textbf{MA} task from dataset-wise mean value predictor (the dashed black line in each graph) and predictors trained with experimental records of other models and~0--5 records from a new model (as indicated by the title of each graph)}.
\label{fig:MA_label}
\end{figure*}
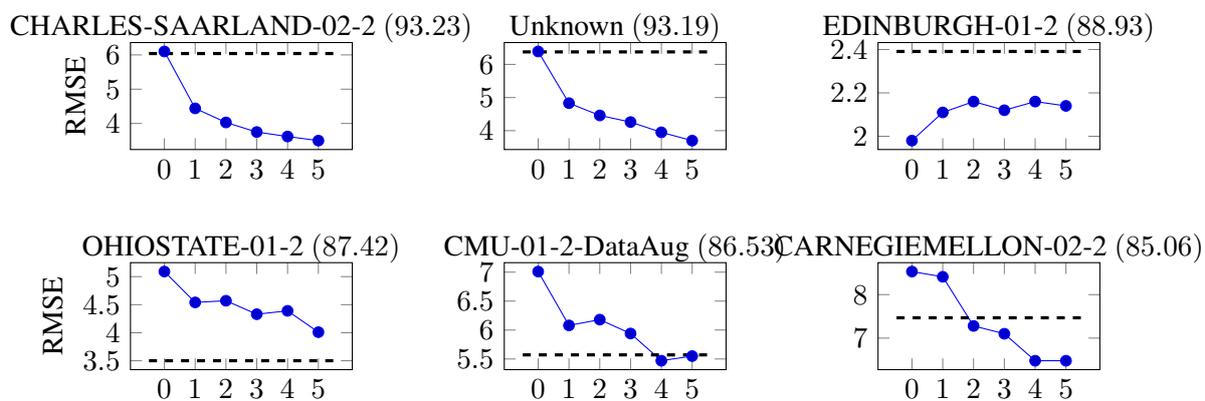

\begin{figure*}[t]
\centering
\pgfplotstableread[row sep=\\,col sep=&]{RMSE & TurkuNLP&CEA&Stanford&Uppsala&AntNLP&ParisNLP&NLP-Cube&SLT-Interactions&IBM&LeisureX&UniMelb&Fudan&KParse&BASELINE\\0 & 18.167434516009777&17.729931167881574&20.140871156154436&15.81372670661391&16.843412042194657&18.33733926746779&18.90846432761132&19.12995469756633&16.25850209588199&17.895641081786067&17.958409470355125&17.63840584483742&16.859293178886578&17.166816745840375\\
1 & 18.07868680500261&18.37411659858157&20.223836433471185&15.536621860319515&16.99277093173855&18.127127190879023&18.650197711993602&19.947854930963437&16.35221157622164&18.039750393724972&17.23314669373558&17.7966805837987&16.75480654018073&17.076931799444328\\
2 & 18.278322937570913&18.33207654677016&20.839309218857895&16.475026275305595&16.513535037681777&18.04748904877911&18.450137113520857&19.48529705890985&16.844916633363248&17.548357373277657&17.759584968477515&18.114617985851&16.283080850613995&16.80402325600139\\
3 & 18.15124493961678&17.707357549462728&20.05881403561267&16.737140521533433&16.96954059788153&18.00357864910683&18.923797377904613&19.74999353654622&16.533917939429415&17.72051934637159&17.400174329308527&17.584266597479772&16.935889150768915&17.316362395065365\\
4 & 18.147435567769485&17.605230994201555&20.79058793540626&16.020595833308715&17.40300016322383&18.709204331263486&18.307717017080513&20.037183828310326&16.141251816185477&18.483187330862524&17.705480939183815&18.031709958580237&16.698772328884232&16.891447664300422\\
5 & 18.254065714618175&18.761979333748737&19.963879970557297&16.469708652470405&17.13630078296084&18.627111512099894&18.517628586531828&20.473250414847964&16.05674926883996&17.64049734706777&17.734071667789156&17.42217663167968&17.226728078001354&17.013640139228823\\
}\basedata
\pgfplotstableread[row sep=\\,col sep=&]{RMSE & TurkuNLP&CEA&Stanford&Uppsala&AntNLP&ParisNLP&NLP-Cube&SLT-Interactions&IBM&LeisureX&UniMelb&Fudan&KParse&BASELINE\\0 & 3.65&2.71&5.38&2.92&2.66&2.57&2.69&6.85&3.49&2.94&2.98&5.2&5.27&5.97\\
1 & 3.08&2.61&5.44&2.81&2.78&2.71&2.82&7.07&3.37&2.6&2.7&4.42&4.28&4.82\\
2 & 3.08&2.52&5.41&2.81&2.77&2.81&2.82&7.12&3.2&2.63&2.55&3.92&3.98&4.32\\
3 & 3.02&2.54&5.39&2.7&2.69&2.75&2.82&7.15&3.12&2.46&2.48&3.88&3.76&4.25\\
4 & 2.96&2.38&5.5&2.61&2.75&2.79&2.65&7.43&3.18&2.41&2.41&3.92&3.68&4.13\\
5 & 2.92&2.51&5.21&2.62&2.76&2.85&2.77&7.88&3.04&2.38&2.42&3.86&3.36&4.1\\
}\systemdata
\pgfplotstableread[row sep=\\,col sep=&]{RMSE & TurkuNLP&CEA&Stanford&Uppsala&AntNLP&ParisNLP&NLP-Cube&SLT-Interactions&IBM&LeisureX&UniMelb&Fudan&KParse&BASELINE\\
-0.5 &4.640813183243175&3.935886706933529&6.299376878325158&3.5533662725602904&3.0521165330500915&3.3193685500674106&3.3369609701042324&7.302848954145549&2.9738454676914947&2.2991094242381793&2.045930717763837&3.987830063046946&3.8203114291206544&4.559047759971615\\
5.5 &4.640813183243175&3.935886706933529&6.299376878325158&3.5533662725602904&3.0521165330500915&3.3193685500674106&3.3369609701042324&7.302848954145549&2.9738454676914947&2.2991094242381793&2.045930717763837&3.987830063046946&3.8203114291206544&4.559047759971615\\
}\basedataa
\begin{tikzpicture}
\begin{groupplot}[group style={group size=4 by 4,vertical sep=1.5cm},height=3cm,width=4.5cm,title style={yshift=-2ex,}]
\nextgroupplot[title=TurkuNLP $(75.93)$,xtick=data,ylabel=RMSE]
\addplot+ table[x=RMSE,y=TurkuNLP]{\systemdata};
\addplot[mark=none, black, very thick, dashed] table[x=RMSE,y=TurkuNLP]{\basedataa};
\nextgroupplot[title=CEA $(75.06)$,xtick=data]
\addplot+ table[x=RMSE,y=CEA]{\systemdata};
\addplot[mark=none, black, very thick, dashed] table[x=RMSE,y=CEA]{\basedataa};
\nextgroupplot[title=Stanford $(75.05)$,xtick=data]
\addplot+ table[x=RMSE,y=Stanford]{\systemdata};
\addplot[mark=none, black, very thick, dashed] table[x=RMSE,y=Stanford]{\basedataa};
\nextgroupplot[title=Uppsala $(74.76)$,xtick=data]
\addplot+ table[x=RMSE,y=Uppsala]{\systemdata};
\addplot[mark=none, black, very thick, dashed] table[x=RMSE,y=Uppsala]{\basedataa};
\nextgroupplot[title=AntNLP $(74.1)$,xtick=data,ylabel=RMSE]
\addplot+ table[x=RMSE,y=AntNLP]{\systemdata};
\addplot[mark=none, black, very thick, dashed] table[x=RMSE,y=AntNLP]{\basedataa};
\nextgroupplot[title=ParisNLP $(74.05)$,xtick=data]
\addplot+ table[x=RMSE,y=ParisNLP]{\systemdata};
\addplot[mark=none, black, very thick, dashed] table[x=RMSE,y=ParisNLP]{\basedataa};
\nextgroupplot[title=NLP-Cube $(73.96)$,xtick=data]
\addplot+ table[x=RMSE,y=NLP-Cube]{\systemdata};
\addplot[mark=none, black, very thick, dashed] table[x=RMSE,y=NLP-Cube]{\basedataa};
\nextgroupplot[title=SLT-Interactions $(72.92)$,xtick=data]
\addplot+ table[x=RMSE,y=SLT-Interactions]{\systemdata};
\addplot[mark=none, black, very thick, dashed] table[x=RMSE,y=SLT-Interactions]{\basedataa};
\nextgroupplot[title=IBM $(71.88)$,xtick=data,ylabel=RMSE]
\addplot+ table[x=RMSE,y=IBM]{\systemdata};
\addplot[mark=none, black, very thick, dashed] table[x=RMSE,y=IBM]{\basedataa};
\nextgroupplot[title=LeisureX $(71.7)$,xtick=data]
\addplot+ table[x=RMSE,y=LeisureX]{\systemdata};
\addplot[mark=none, black, very thick, dashed] table[x=RMSE,y=LeisureX]{\basedataa};
\nextgroupplot[title=UniMelb $(71.54)$,xtick=data]
\addplot+ table[x=RMSE,y=UniMelb]{\systemdata};
\addplot[mark=none, black, very thick, dashed] table[x=RMSE,y=UniMelb]{\basedataa};
\nextgroupplot[title=Fudan $(69.42)$,xtick=data]
\addplot+ table[x=RMSE,y=Fudan]{\systemdata};
\addplot[mark=none, black, very thick, dashed] table[x=RMSE,y=Fudan]{\basedataa};
\nextgroupplot[title=KParse $(69.39)$,xtick=data,ylabel=RMSE]
\addplot+ table[x=RMSE,y=KParse]{\systemdata};
\addplot[mark=none, black, very thick, dashed] table[x=RMSE,y=KParse]{\basedataa};
\nextgroupplot[title=BASELINE $(68.5)$,xtick=data]
\addplot+ table[x=RMSE,y=BASELINE]{\systemdata};
\addplot[mark=none, black, very thick, dashed] table[x=RMSE,y=BASELINE]{\basedataa};
\end{groupplot}
\end{tikzpicture}
\caption{RMSE scores of \textbf{UD} task from dataset-wise mean value predictor (the dashed black line in each graph) and predictors trained with experimental records of other models and~0--5 records from a new model (as indicated by the title of each graph).}
\label{fig:UD_label2}
\end{figure*}
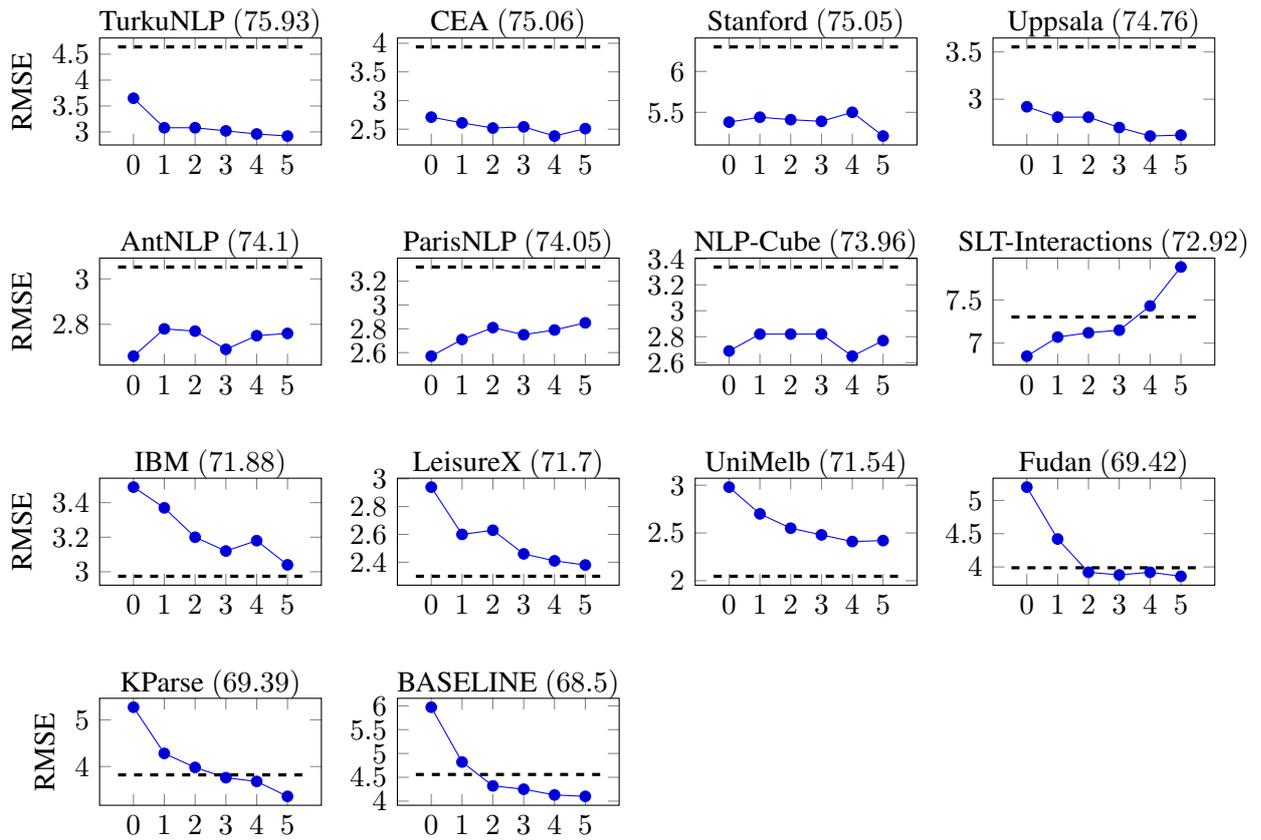
\clearpage
\newpage
\section{Feature importance}
In this section, we show the plots of feature importance for all the tasks.

\begin{figure}[h!]
\centering
  \begin{subfigure}{\linewidth}
  \includegraphics[width=\linewidth]{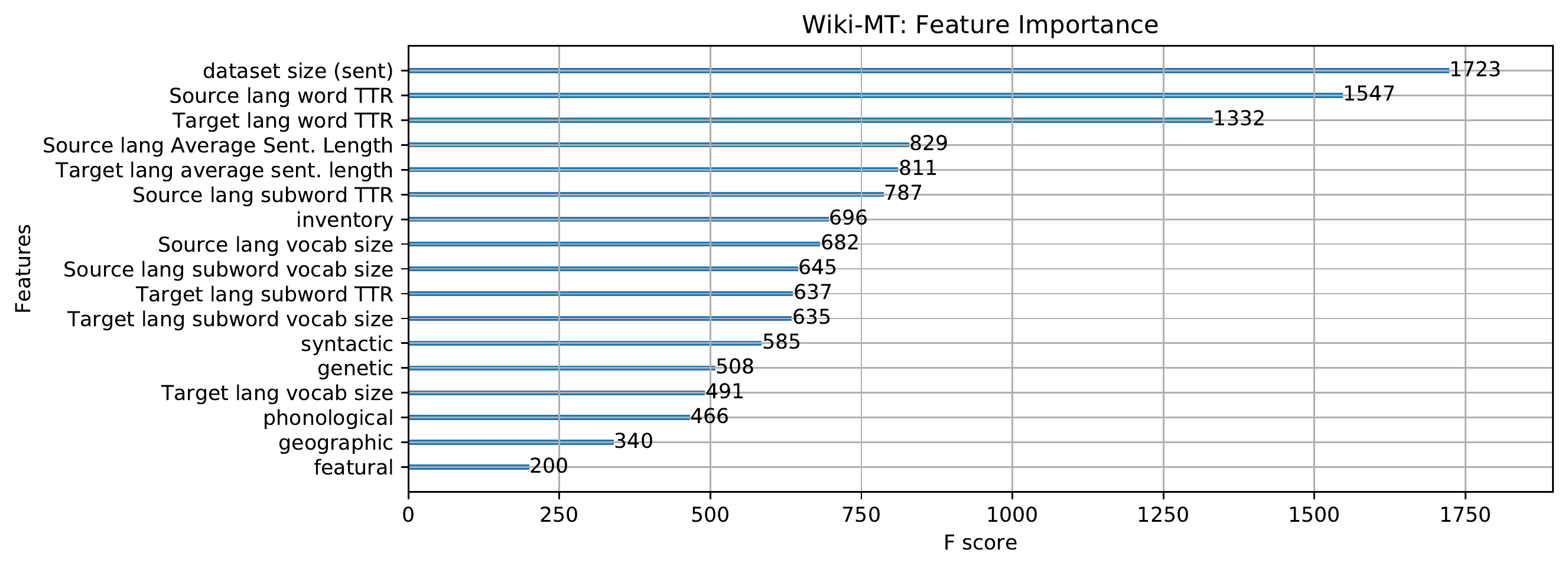}
  
  \end{subfigure}\par\medskip
\end{figure}

\begin{figure}[h!]
  \ContinuedFloat
  \begin{subfigure}{\linewidth}
  \includegraphics[width=\linewidth]{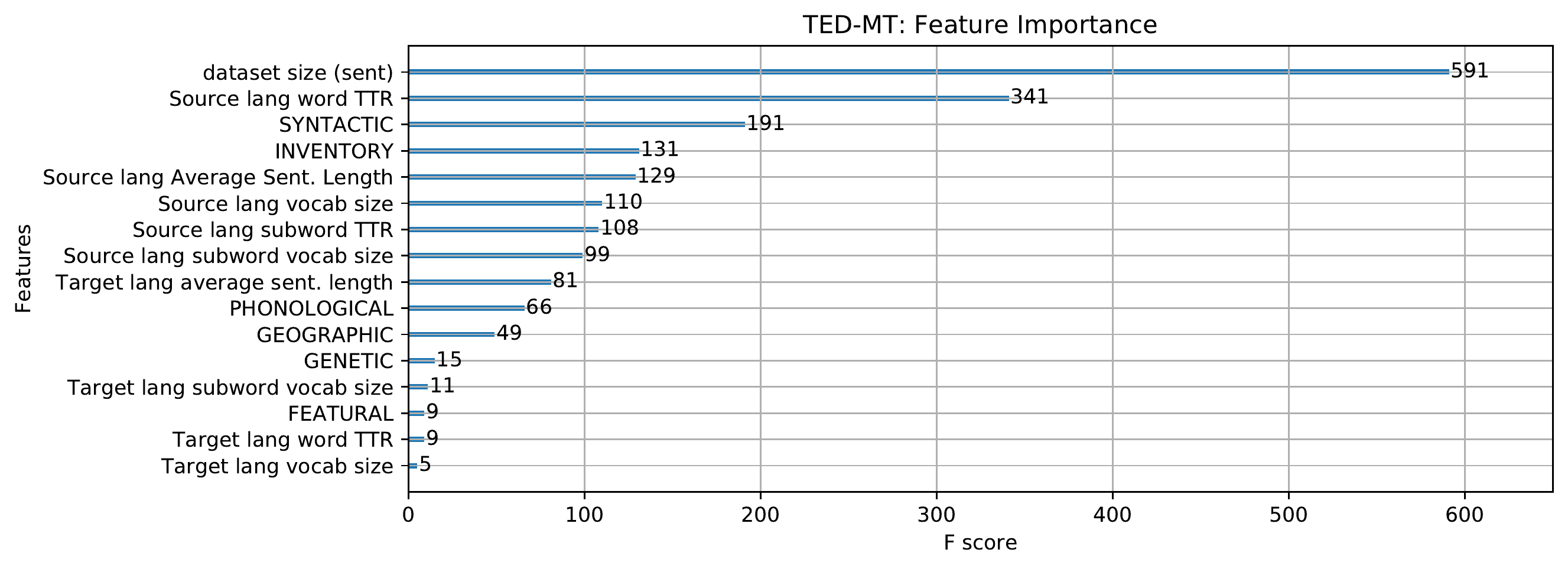}
  
  \end{subfigure}\par\medskip
\end{figure}

\begin{figure}[h!]
  \ContinuedFloat
  \begin{subfigure}{\linewidth}
  \includegraphics[width=\linewidth]{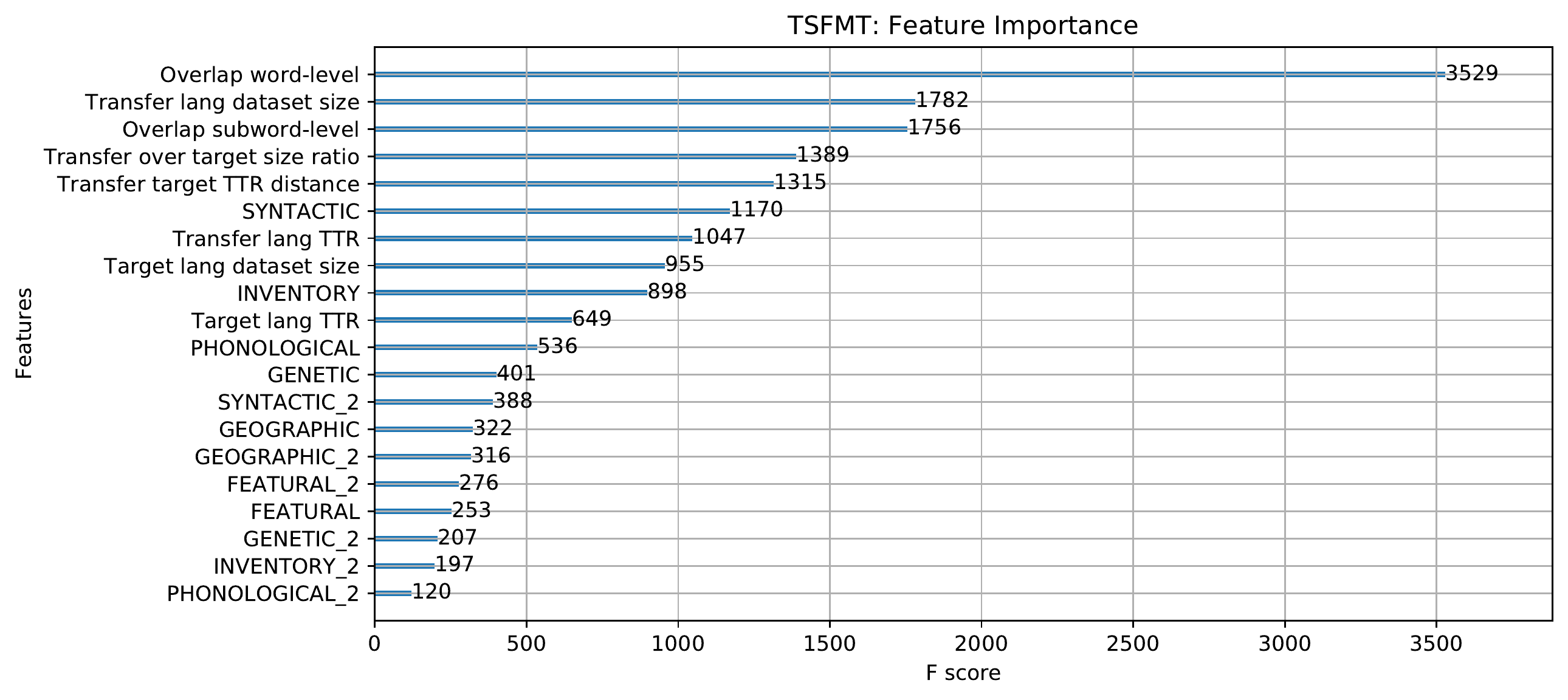}
  
  \end{subfigure}\par\medskip
\end{figure}

\begin{figure}[h!]
  \ContinuedFloat
  \begin{subfigure}{\linewidth}
  \includegraphics[width=\linewidth]{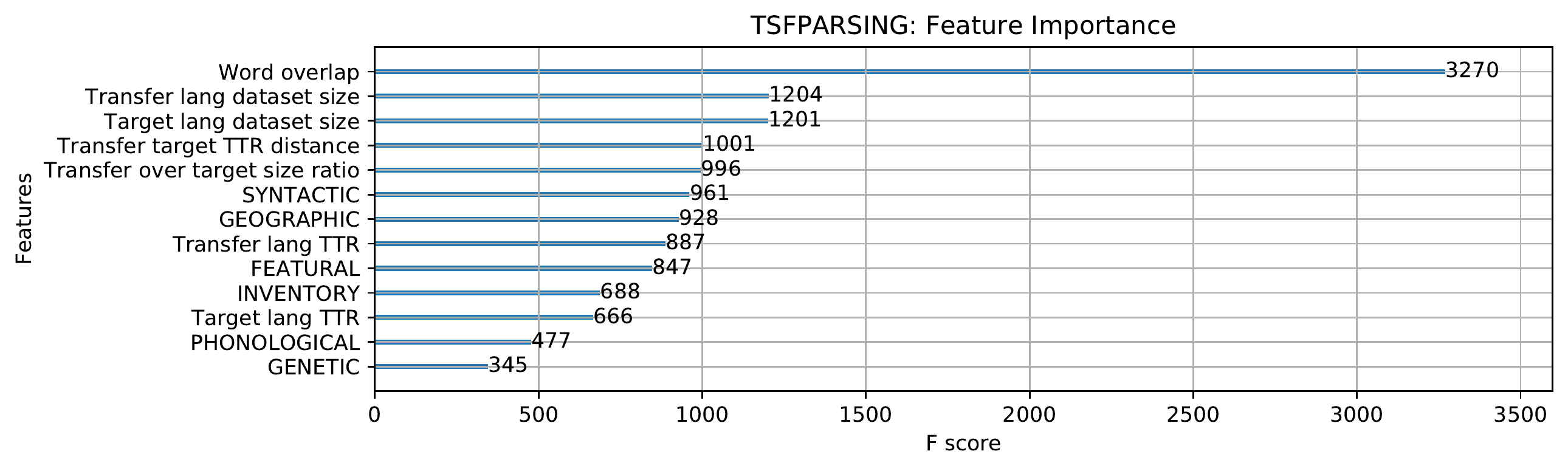}
  
  \end{subfigure}\par\medskip
\end{figure}

\begin{figure}[h!]
  \ContinuedFloat
  \begin{subfigure}{\linewidth}
  \includegraphics[width=\linewidth]{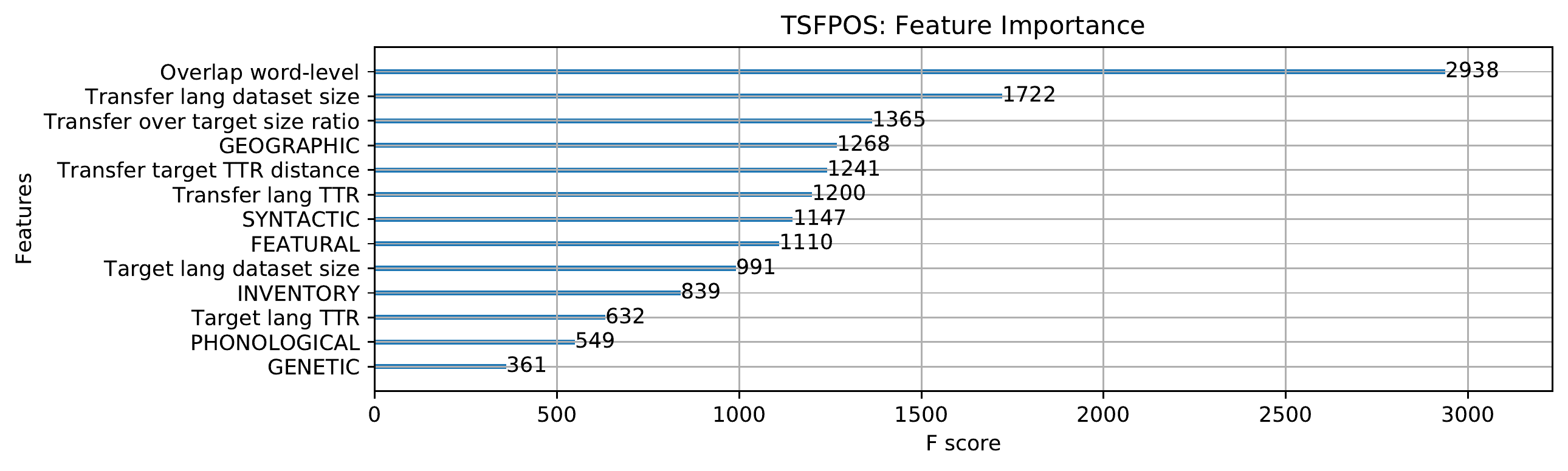}
  
  \end{subfigure}\par\medskip
\end{figure}

\begin{figure}[h!]
  \ContinuedFloat
  \begin{subfigure}{\linewidth}
  \includegraphics[width=\linewidth]{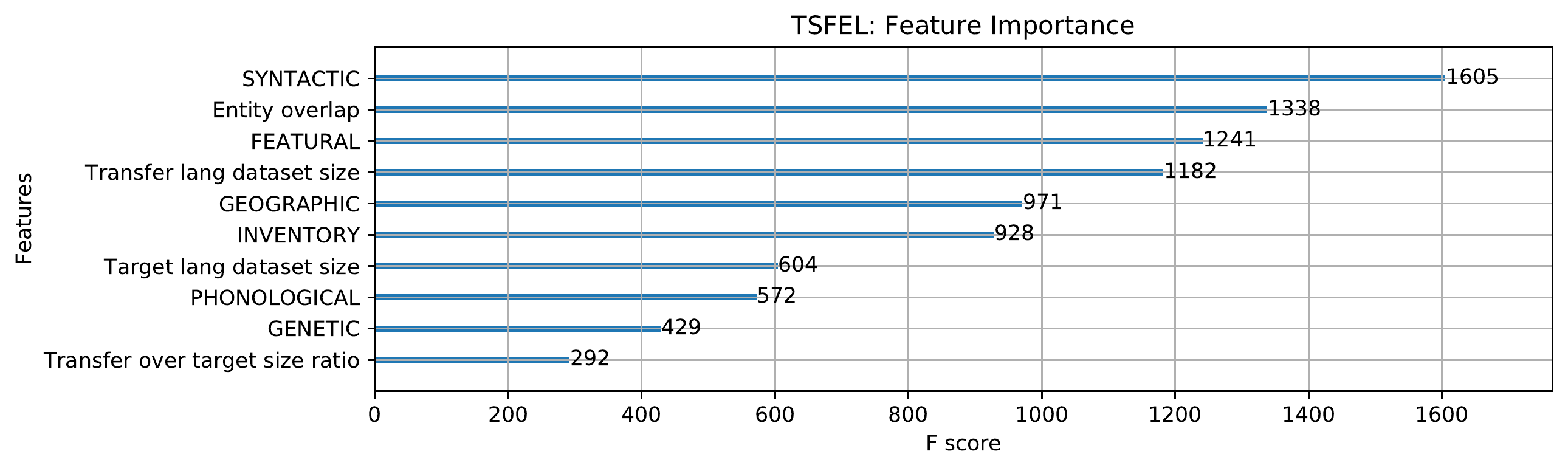}
  
  \end{subfigure}
\end{figure}

\begin{figure}[h!]
  \ContinuedFloat
  \centering
  \begin{subfigure}{0.9\linewidth}
  \includegraphics[width=\linewidth]{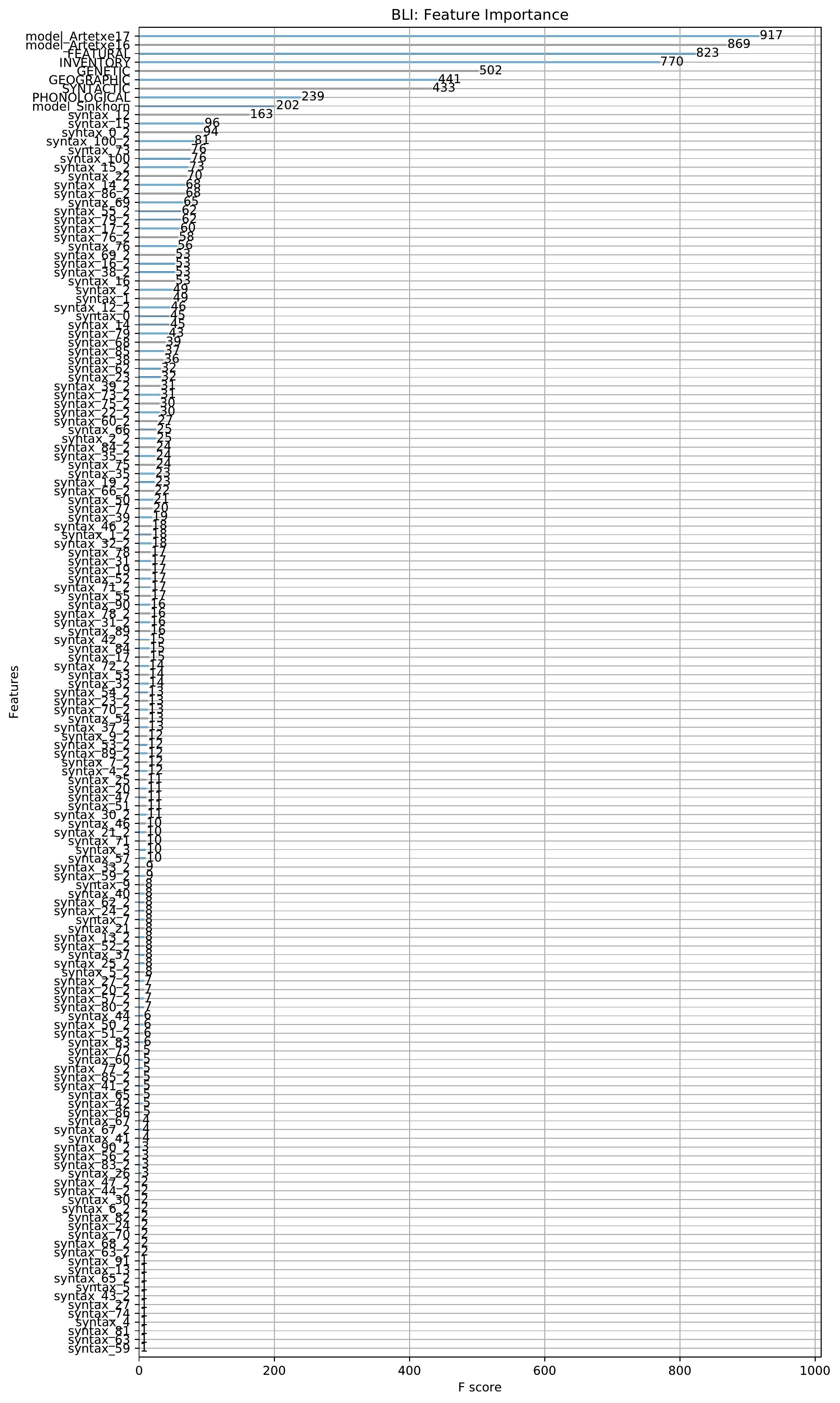}
  
  \end{subfigure}
\end{figure}

\begin{figure}[h!]
  \ContinuedFloat
  \begin{subfigure}{\linewidth}
  \includegraphics[width=\linewidth]{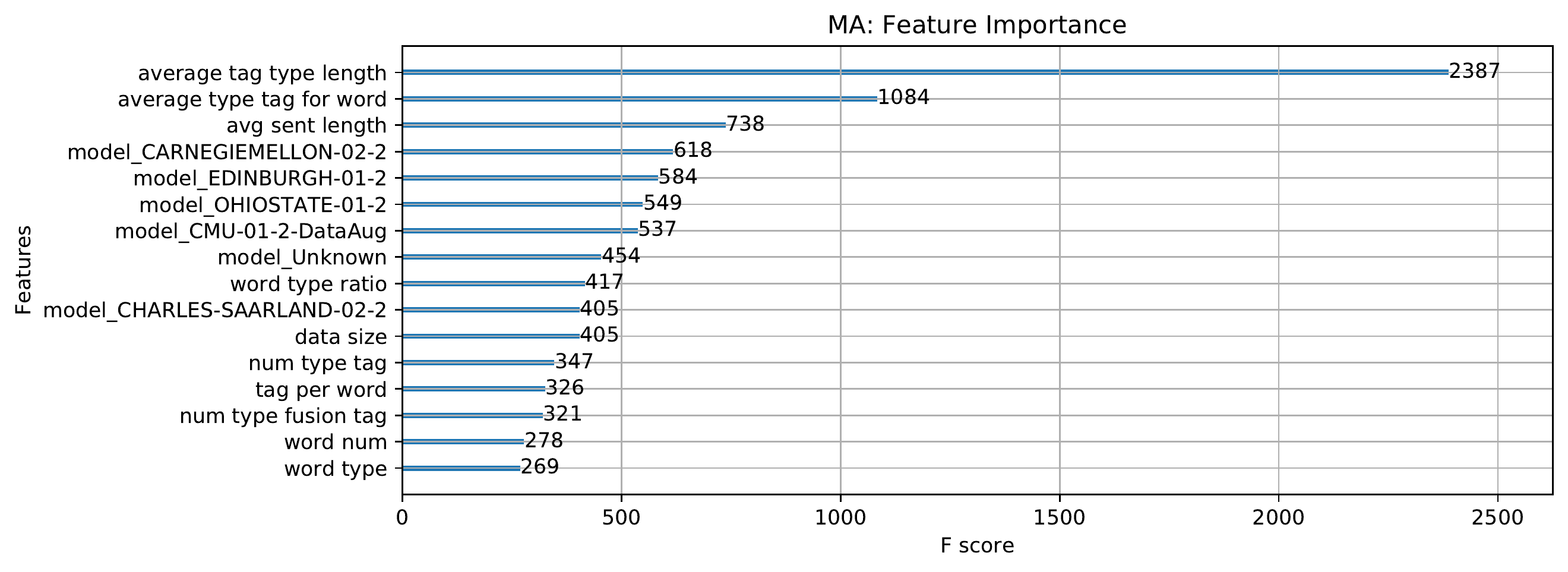}
  
  \end{subfigure}\par\medskip
\end{figure}

\begin{figure}[h!]
  \ContinuedFloat
  \begin{subfigure}{\linewidth}
  \includegraphics[width=\linewidth]{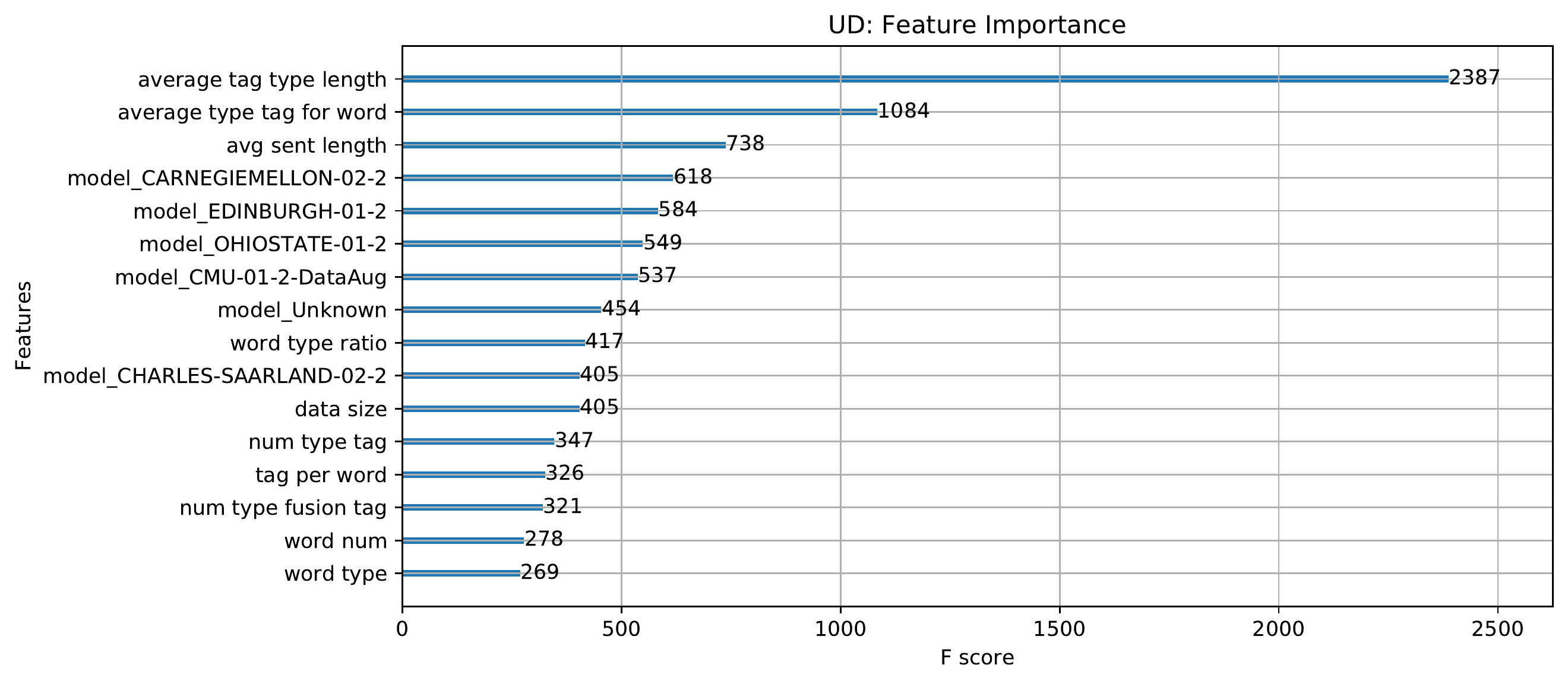}
  
  \end{subfigure}
\end{figure}

\end{document}